\documentclass[runningheads]{llncs}
\usepackage{comment} 
\usepackage{color}

\usepackage{epsfig}
\usepackage{graphicx}
\usepackage{amsmath}
\usepackage{amssymb}
\usepackage{multirow}
\usepackage{makecell}
\usepackage{booktabs}
\usepackage{subfigure}

\begin{document}
\pagestyle{headings}
\mainmatter

\makeatletter
\renewcommand*{\@fnsymbol}[1]{\ensuremath{\ifcase#1\or \dagger \or * \or \ddagger\or
    \mathsection\or \mathparagraph\or \|\or **\or \dagger\dagger
    \or \ddagger\ddagger \else\@ctrerr\fi}}
\makeatother

\title{Negative Margin Matters: Understanding Margin in Few-shot Classification}
\author{Bin Liu\inst{1}\thanks{Equal contribution. $^*$ The work is done when Yutong Lin is an intern at MSRA.} \and
Yue Cao\inst{2\dagger} \and
Yutong Lin\inst{2,3} \and
Qi Li\inst{1} \and
Zheng Zhang\inst{2} \and \\
Mingsheng Long\inst{1} \and
Han Hu\inst{2}
}
\authorrunning{Liu et al.}
\institute{$^1$Tsinghua University ~~~
$^2$Microsoft Research Asia ~~~
$^3$Xi'an Jiaotong University}

\maketitle
\begin{abstract}
This paper introduces a negative margin loss to metric learning based few-shot learning methods. The negative margin loss significantly outperforms regular softmax loss, and achieves state-of-the-art accuracy on three standard few-shot classification benchmarks with few bells and whistles. 
These results are contrary to the common practice in the metric learning field, that the margin is zero or positive.
To understand why the negative margin loss performs well for the few-shot classification, we analyze the discriminability of learned features w.r.t different margins for training and novel classes, both empirically and theoretically. We find that although negative margin reduces the feature discriminability for training classes, it may also avoid falsely mapping samples of the same novel class to multiple peaks or clusters, and thus benefit the discrimination of novel classes. Code is available at \url{https://github.com/bl0/negative-margin.few-shot}.
\vspace{-1em}

\end{abstract}

\section{Introduction}

Recent success on visual recognition tasks \cite{krizhevsky2012alexnet,simonyan2014very,he2015resnet,ren2015fasterrcnn,dai2017dcnv1,carreira2017i3d} heavily relies on the massive-scale manually labeled training data, which is too expensive in many real scenarios.
In contrast, humans are capable of learning new concepts with only a few examples, yet it still remains a challenge for modern machine learning systems.
Hence, learning to generalize the knowledge in base classes (with sufficient annotated examples) to novel classes (with a few labeled examples), also known as few-shot learning, has attracted more and more attention \cite{chen2019closerfewshot,lee2019metaopt,mangla2019charting,dhillon2019baseline,vinyals2016matching,snell2017prototypical,finn2017MAML,ravi2016optimization,sung2018relation,qiao2018few,qi2018low,garcia2017few}.

\begin{figure}
    \centering
    \subfigure[1-shot accuracy]{
        \includegraphics[width=0.40\linewidth]{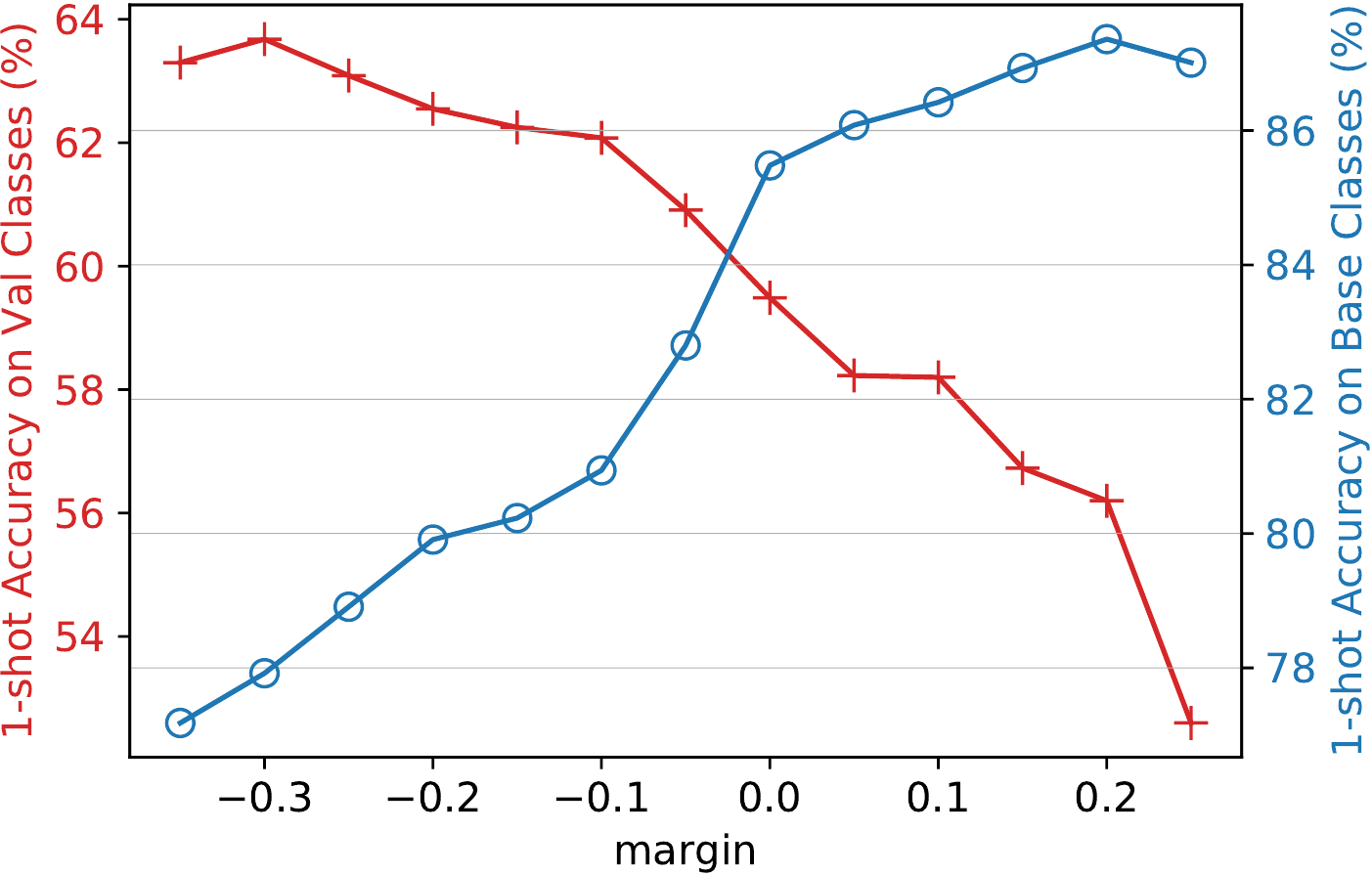}
    }
    \subfigure[5-shot accuracy]{
        \includegraphics[width=0.43\linewidth]{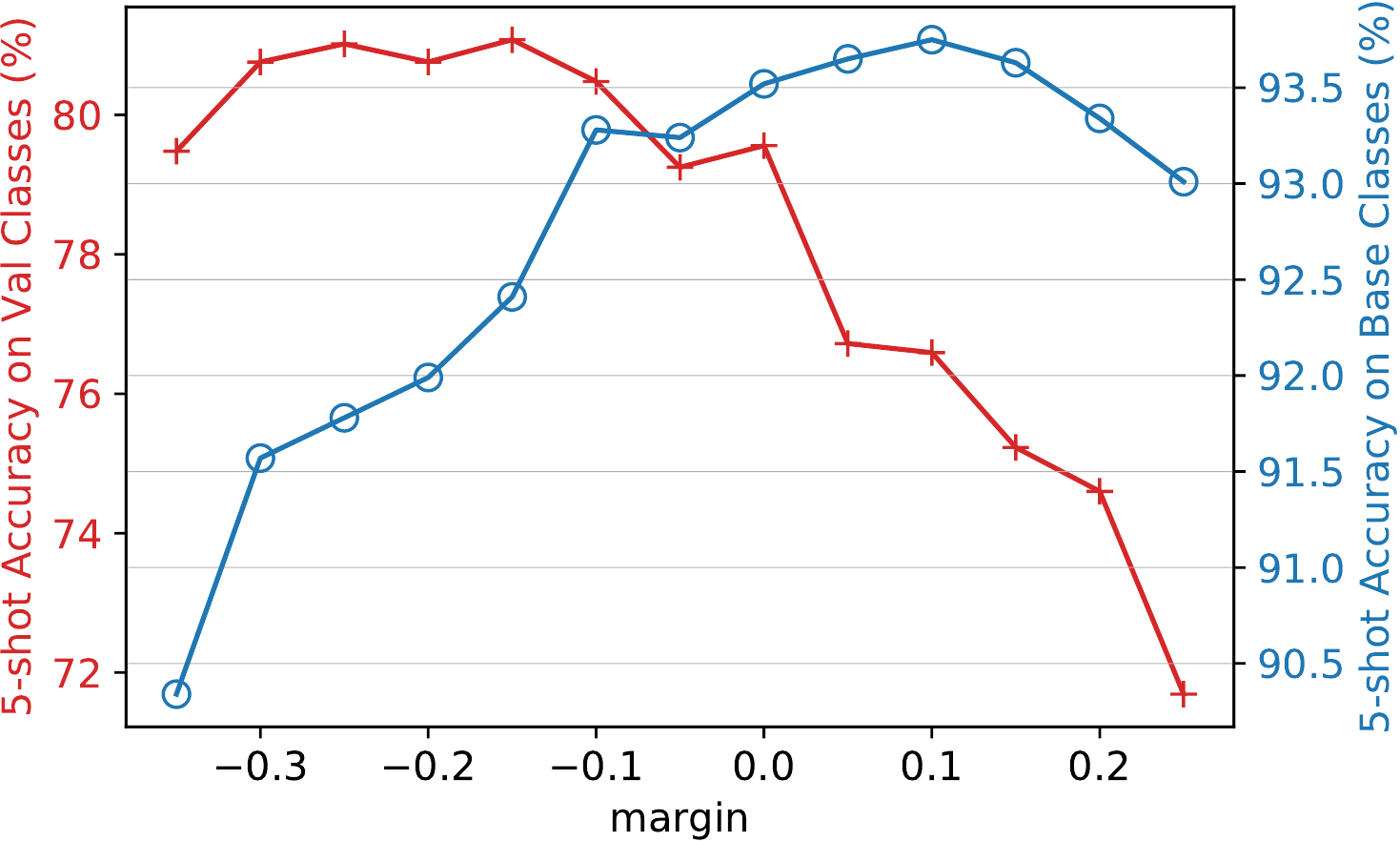}
    }
    \vspace{-5pt}
    \caption{The one-shot and five-shot accuracy on novel classes (in red) and base classes (in blue) w.r.t different margins in cosine softmax loss on mini-ImageNet. As we expect, applying larger margin to softmax loss can achieve better accuracy on base classes. But surprisingly, applying appropriate negative margin to softmax loss can achieve state-of-the-art few-shot accuracy on novel classes.}
    \label{fig:teaser}
\vspace{-5pt}
\end{figure}

An important direction of few-shot classification is meta learning, which aims to learn a meta-learner on base classes and generalizes it to novel classes. Metric learning based methods \cite{chen2019closerfewshot,dhillon2019baseline,mangla2019charting}, are an important series of the meta-learning methods, and perform metric learning in the base classes and then transfer the learned metrics to the novel classes.
For example, \cite{chen2019closerfewshot} proved that simply using standard softmax loss or cosine softmax loss for learning metrics in base classes can achieve the state-of-the-art few-shot classification performance via learning a linear classifier on novel classes.

In the metric learning area, a common view is that the standard softmax loss is insufficient for discrimination on different training classes. Several previous approaches integrate the large and positive margin to the softmax loss \cite{liu2016largesoftmax} or the cosine softmax loss \cite{deng2019arcface,wang2018cosface} so as to enforce the score of ground truth class larger than that of other classes by at least a margin. This could help to learn highly-discriminative deep features and result in remarkable performance improvement on visual recognition tasks, especially on face recognition \cite{liu2016largesoftmax,deng2019arcface,wang2018cosface}. 

Consequently, it inspires us to adopt this large-margin softmax loss to learn better metrics for few-shot classification. 
As we expected, shown as the blue curves in Fig.~\ref{fig:teaser}, the metrics learned by large-margin softmax with positive margin are more discriminative on training classes, resulting in higher few-shot accuracy on the validation set of training classes.
But in the standard open-set setting of few-shot classification, shown as red curves in Fig.~\ref{fig:teaser}, we surprisingly find out that adding the positive margin in softmax loss would hurt the performance.

From our perspective, the positive margin would make the learned metrics more discriminative to training classes. But for novel classes, positive margin would map the samples of the same class to multiple peaks or clusters in base classes (shown in Fig.~\ref{fig:mnist} and Fig.~\ref{fig:tsne}) and hurt their discriminability. 
We then give a theoretical analysis that the discriminability of the samples in the novel classes is monotonic decreasing w.r.t the margin parameter under proper assumption. 
Instead, appropriate negative margin could achieve a better tradeoff between the discriminability and transferability for novel classes, and achieves better performance on few-shot classification.

The main contributions of this paper are summarized as follows:
\begin{enumerate}
    \item This is the first endeavor to show that softmax loss with negative margin works surprisingly well on few-shot classification, which breaks the inherent understanding that margin can only be limited to positive values \cite{deng2019arcface,liu2016largesoftmax,wang2018cosface}.
    \item We provide insightful intuitive explanation and the theoretical analysis about why negative margin works well for few-shot classification.
    \item The proposed approach with negative margin achieves state-of-the-art performance on three widely-used few-shot classification benchmarks.
\end{enumerate}

\section{Related Work}
\textbf{Few-Shot Classification.}
The existing representative few-shot learning methods can be broadly divided into three categories: \textit{gradient-based} methods, \textit{hallucination}\textit{-based} methods, and \textit{metric-based} methods.

\textit{Gradient-based} methods tackle the few-shot classification by learning the task-agnostic knowledge. 
\cite{finn2017MAML,rusu2018LEO,nichol2018first,ICML18MunkhdalaiRapid,mishra2017simple} focus on learning a suitable initialization of the model parameters which can quickly adapt to new tasks with a limited number of labeled data and a small number of gradient update steps.
Another line of works aims at learning an optimizer, such as LSTM-based meta learner \cite{ravi2016optimization} and weight-update mechanism with an external memory \cite{munkhdalai2017meta},  for replacing the stochastic gradient descent optimizer.
However, it is challenging to solve the dual or bi-level optimization problem of these works, so their performance is not competitive on large datasets.
Recently, \cite{lee2019metaopt,bertinetto2018R2D2} alleviate the optimization problem by closed-form model like SVM, and achieve better performance on few-shot classification benchmark of large dataset.

\textit{Hallucination-based} methods attempt to address the limited data issue by learning an image generator from base classes, which is adopted to hallucinate new images in novel classes \cite{hariharan2017low,wang2018low}. 
\cite{hariharan2017low} presents a way of hallucinating additional examples for novel classes by transferring modes of variation from base classes. 
\cite{wang2018low} learns to hallucinate examples that are useful for classification by the end-to-end optimization of both classifier and hallucinator.
As hallucination-based methods can be considered as the supplement and are always adopted with other few-shot methods, we follow \cite{chen2019closerfewshot} to exclude these methods in our experimental comparison and leave it to future work.

\textit{Metric-based} methods aim at learning a transferable distance metric. 
MatchingNet~\cite{vinyals2016matching} computes cosine similarity between the embeddings of labeled images and unlabeled images, to classify the unlabeled images.
ProtoNet~\cite{snell2017prototypical} represents each class by the mean embedding of the examples inside this class, and the classification is performed based on the distance to the mean embedding of each class.
RelationNet~\cite{sung2018relation} replaces the non-parametric distance in ProtoNet to a parametric relation module.
Recently, \cite{chen2019closerfewshot,dhillon2019baseline,mangla2019charting} reveal that the simple pre-training and fine-tuning pipeline (following the standard transfer learning paradigm) can achieve surprisingly competitive performance with the state-of-the-art few-shot classification methods.

Based on this simple paradigm, our work is the first endeavor towards explicitly integrating the margin parameter to the softmax loss, and mostly importantly breaks the inherent understanding that the margin can be only restricted as positive values, with both intuitive understanding and theoretical analysis. With an appropriate negative margin, our approach could achieve the state-of-the-art performance on three standard few-shot classification benchmarks.

\textbf{Margin based Metric Learning.} 
Metric learning aims to learn a distance metric between examples, and plays a critical role in many tasks, such as classification~\cite{weinberger2009distance}, clustering~\cite{xing2003distance}, retrieval~\cite{liu2018DTQ} and visualization~\cite{cite:tsne}.

In practice, the margin between data points and the decision boundary plays a significant role in achieving strong generalization performance. \cite{koltchinskii2002empirical} develops a margin theory and shows that the margin loss leads to an informative generalization bound for classification task.
In the past decades, the idea of margin-based metric learning has been widely explored in SVM~\cite{scholkopf2002learning}, k-NN classification~\cite{weinberger2009distance}, multi-task learning~\cite{parameswaran2010large}, etc.
In the deep learning era, many margin-based metric learning methods are proposed to enhance the discriminative power of the learned deep features, and show remarkable performance improvements in many tasks~\cite{liu2018DTQ,lee2019metaopt,narayanaswamy2019designing}, especially in face verification~\cite{schroff2015facenet,liu2017sphereface,wang2018cosface,deng2019arcface}. 
For example, SphereFace~\cite{liu2017sphereface},
CosFace~\cite{wang2018cosface}, and
ArcFace~\cite{deng2019arcface} enforce the intra-class variance and inter-class diversity by adding the margin to cosine softmax loss.

However, as the tasks of previous works are based on close-set scenarios, they limit the margin parameter as positive values \cite{liu2017sphereface,wang2018cosface,deng2019arcface}, where making the deep features more discriminative could be generalized to the validation set and improve the performance.
For open-set scenarios, such as few-shot learning, increasing the margin would not enforce the inter-class diversity but unfortunately enlarge the intra-class variance for novel classes, as shown in Fig.~\ref{fig:discriminative_function}, which would hurt the performance. In contrast, an appropriate negative margin would better tradeoff the discriminability and transferability of deep features in novel classes, and obtain better performance for few-shot classification.


\section{Methodology}
In a few-shot classification task, we are given two sets of data with different classes, formulated as $I^b=\{({\bf x}_i, y_i)\}_{i=1}^{N^b}$ as the base training set with $C^b$ base classes for the first training stage, and $I^n=\{({\bf x}_i', y_i')\}_{i=1}^{N^n}$ as the novel training set with $C^n$ novel classes for the second training stage. 
For the novel training set, each class has $K$ samples, where $K=1$ or $5$, and $C^n = 5$ is the standard setting \cite{chen2019closerfewshot,lee2019metaopt,mangla2019charting,dhillon2019baseline,sung2018relation,qiao2018few,qi2018low}.
This is called $C^n$-way $K$-shot learning.
Few-shot classification aims to learn both discriminative and transferable feature representations from the abundant labeled data in base classes, such that the features can be easily adapted for the novel classes with few labeled examples. 

\subsection{Negative-margin Softmax Loss}\label{sec:softmax}
In image classification, the softmax loss is built upon the feature representation of deep networks ${\bf z}_i=f_{\theta}({\bf x}_i) \in \mathbb{R}^D$ ($f_{\theta}(\cdot)$ denotes the backbone network with the parameters $\theta$), its corresponding label $y_i$ and the linear transform matrix ${\bf W}=[W_1, W_2, ..., W_{C^b}]\in \mathbb{R}^{D\times C^b}$.
Recently, introducing the \textbf{large and positive} margin parameter to the softmax loss is widely explored in metric learning \cite{liu2016largesoftmax,wang2018cosface,deng2019arcface}.
Hence, we directly integrate the margin parameter to the softmax loss to learn the transferable metrics, aiming at benefiting the few-shot classification on novel classes.
The general formulation of large-margin softmax loss is defined as
\begin{equation}\label{eqn:nm_softmax_general}
    \vspace{-5pt}
    L =  - \frac{1}{N}\sum\limits_{i = 1}^N {\log \frac{{{e^{\beta  \cdot \left( {\mathtt{s}\left( {{z_i},{W_{{y_i}}}} \right) - m} \right)}}}}{{{e^{\beta  \cdot \left( {\mathtt{s}\left( {{z_i},{W_{{y_i}}}} \right) - m} \right)}} + \sum\limits_{j = 1,j \ne {y_i}}^C {{e^{\beta  \cdot \mathtt{s}\left( {{z_i},{W_j}} \right)}}} }}} ,
\end{equation}
where $m$ is the margin parameter, $\beta$ denotes the temperature parameter which could help to apply the appropriate concentrated level around the largest scores. And $\mathtt{s}(\cdot,\cdot)$ denotes the similarity function between two input vectors.

It's worth noting that all the previous works on large-margin softmax loss restrict the margin as positive values 
\cite{liu2016largesoftmax,wang2018cosface,deng2019arcface}.
This is because that previous works focus on the close-set scenarios, the loss with larger margin leads to the smaller intra-class variance and the larger between-class variance, which will help to classify examples in the same classes.
This is also validated in Figure~\ref{fig:teaser}, that the softmax loss with larger margin could improve the classification accuracy on the validation set of training classes.

However, the situations are different in the open-set scenarios. Learned metrics which are too discriminative to training classes may hurt their transferability to the novel classes. So applying appropriate negative margin to softmax loss aims to tradeoff the discriminability on training classes and the transferability to novel classes of the learned metrics.

Here we formulate two instantiations of Eqn.~\ref{eqn:nm_softmax_general} with different similarity functions. By taking the inner-product similarity $\mathtt{s}\left( {{{\bf{z}}_i},{W_j}} \right) = W_j^T{{\bf{z}}_i}$ into Eqn.~\ref{eqn:nm_softmax_general}, the \textbf{negative-margin softmax loss} (abbreviated as Neg-Softmax) could be obtained.
By taking the cosine similarity $\mathtt{s}\left( {{{\bf{z}}_i},{W_j}} \right) = \frac{{W_j^T{{\bf{z}}_i}}}{{\left\| {{{\bf{z}}_i}} \right\|\left\| {{W_j}} \right\|}}$ into Eqn.~\ref{eqn:nm_softmax_general}, we can formulate the \textbf{negative-margin cosine softmax loss} (abbreviated as Neg-Cosine). The detailed loss functions could be found at the Appendix. These two loss functions are adopted at the pre-training stage, as shown in Figure~\ref{fig:arch}.

\subsection{Discriminability analysis of deep features w.r.t different margins}

We analyze the discriminability of the deep features extracted by the deep model with different margins, to understand why negative margin works well on novel classes.
For simplicity, we only analyze the cosine softmax loss, and it is direct to extend the analysis and conclusion to standard softmax loss.

We denote the pre-trained backbone network trained with margin parameter $m$ as $f_{\theta(m)}$. 
For class $j$ in base classes or novel classes, denote the set of examples labeled with class $j$ as $I_j = \{(x_i, y_i) | y_i = j\}$.
We compute the class center $\mu(I_j, m)$ for class $j$ as the mean of the L2-normalized feature embeddings as
\begin{equation}
    \vspace{-5pt}
    \begin{aligned}
        \mu(I_j, m)
        = \frac{1}{\left|I_{j}\right|} \sum_{(\mathbf{x}_{i}, y_i) \in I_{j}} \frac{f_{\theta(m)}\left(\mathbf{x}_{i}\right)}{\| f_{\theta(m)}\left(\mathbf{x}_{i}\right) \|_2} .
    \end{aligned}
\end{equation}
The dataset $I = I_1 \cup I_2 \cup \cdots \cup I_C$ with C classes could be base dataset $I^b$ with a large number of base classes or novel dataset $I^n$ with small number of novel classes (such as 5 for 5-way few shot learning). Then we define the inter-class variance $D_{\text{inter}}(I, m)$, and intra-class variance $D_{\text{intra}}(I, m)$ as 
\begin{equation}
\footnotesize
    \vspace{-5pt}
    \begin{aligned}
        D_{\text{inter}}(I, m) &= \frac{1}{C (C - 1)} \sum_{j=1}^C \sum_{k=1, k\neq j}^C  \| \mu(I_j, m) - \mu(I_k, m)  \|_2^2 , \\
        D_{\text{intra}}(I, m) &= \frac{1}{C} \sum_{j=1}^{C} 
        ( 
            \frac{1}{|I_j|} \sum_{(\mathbf{x}_{i}, y_i) \in I_j} \left\Vert \frac{f_{\theta(m)}(x_i)}{\| f_{\theta(m)}(x_i) \|} - \mu(I_j, m) \right\Vert_2^2
        )
        .
    \end{aligned}
    \normalsize
\end{equation}
For every two classes, the inter-class variance is the squared L2 distance between their class centers. 
For each class, the intra-class variance is the squared L2 distance between every sample in this class and the class center.

\begin{figure*}[tpb]
    \centering
    \subfigure[$D_{\text{inter}}$]{
        \includegraphics[width=0.3\textwidth]{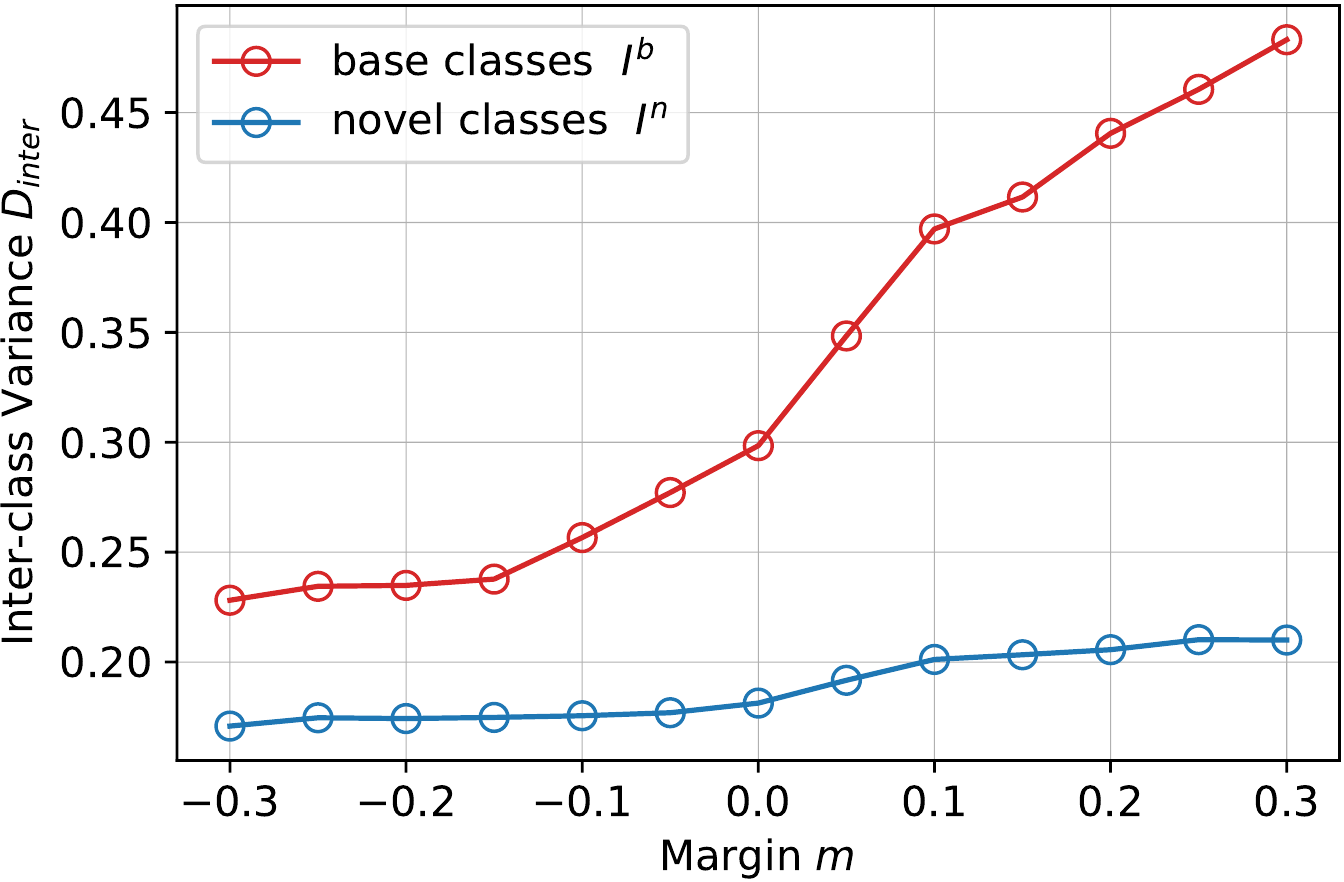}
    }
    \hfill
    \subfigure[$D_{\text{intra}}$]{
        \includegraphics[width=0.3\textwidth]{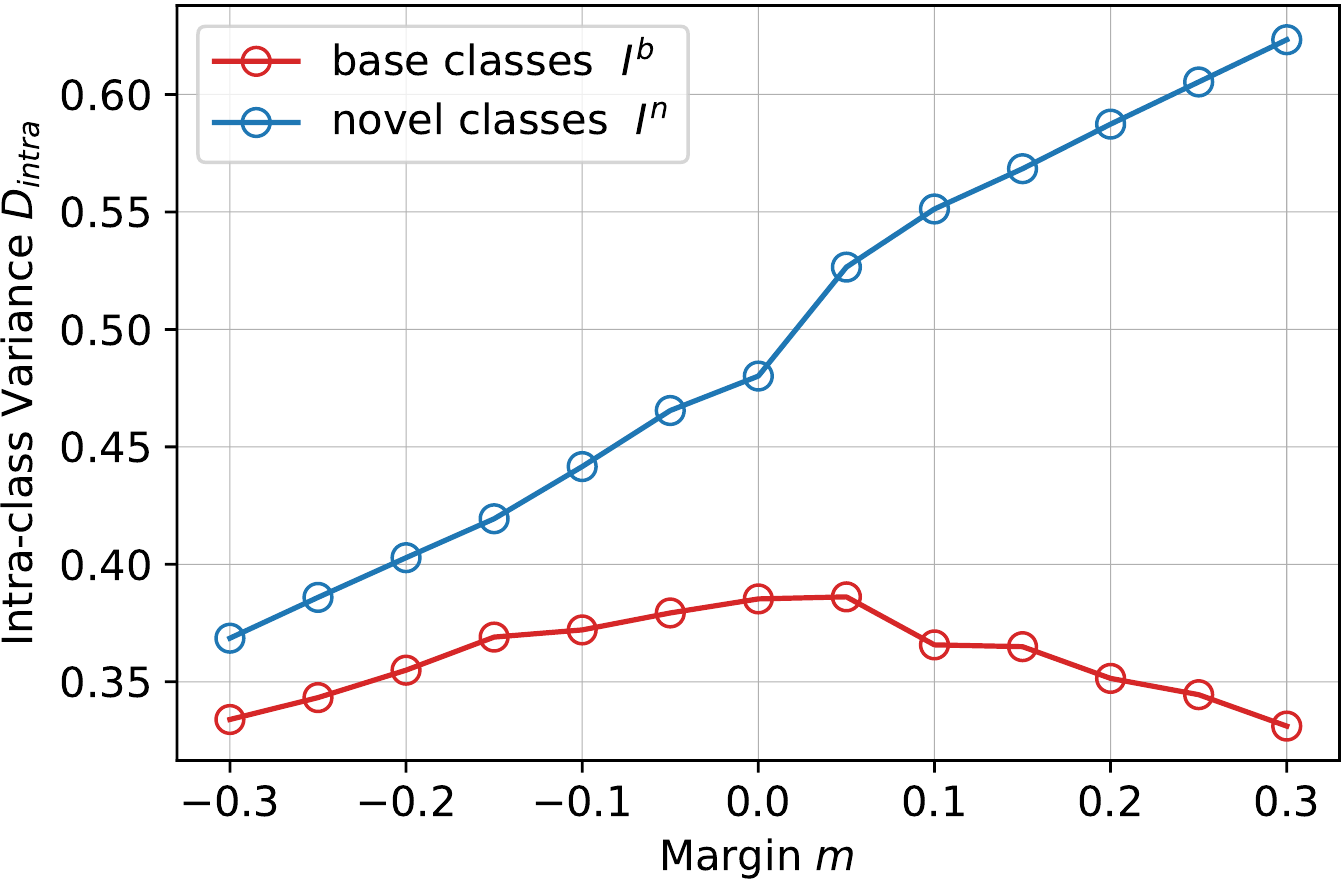}
    }
    \hfill
    \subfigure[$\phi$]{
        \includegraphics[width=0.3\textwidth]{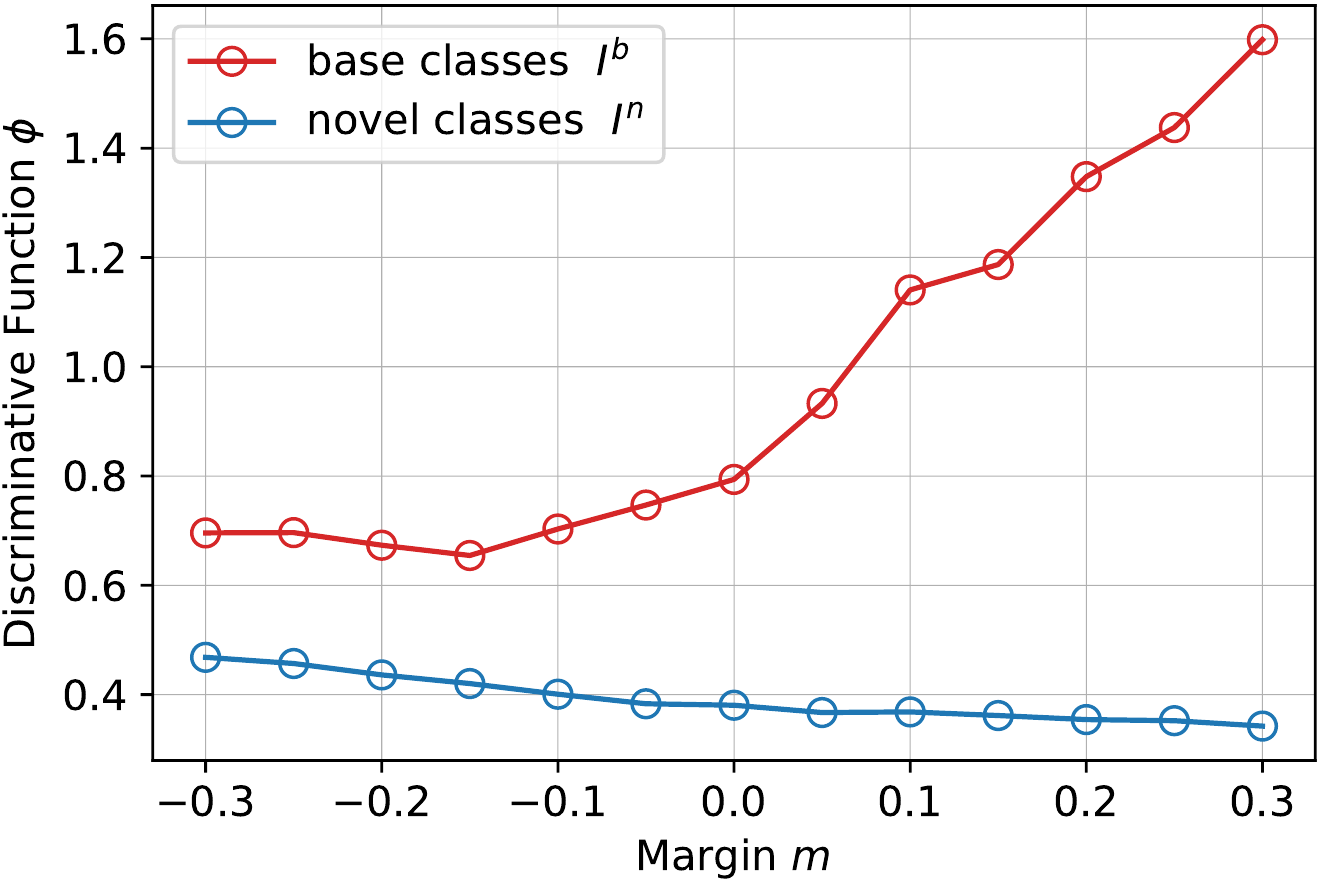}
    }
    \vspace{-15pt}
    \caption{Inter-class variance $D_{\text{inter}}$, intra-class variance $D_{\text{intra}}$, and discriminative function $\phi$ w.r.t margin $m$ on both base and novel classes of mini-ImageNet. As the margin increases, the features of base classes is more discriminative, while that of novel classes is less discriminative.}
    \label{fig:discriminative_function}
    \vspace{-5pt}
\end{figure*}

If inter-class variance becomes larger or intra-class variance becomes smaller, the deep features would be more discriminative.
So we follow \cite{mika1999fisher} to define the discriminative function $\phi(I, m)$ as the inter-class variance divided by the intra-class variance:
\begin{equation}
    \vspace{-5pt}
    \phi(I, m) = \frac{D_{\text{inter}}(I, m)}{D_{\text{intra}}(I, m)}  .
\end{equation}
To measure the discriminability of the deep features with different margins, we plot the inter-class variance $D_{\text{inter}}$, intra-class variance $D_{\text{intra}}$, and discriminative function $\phi$ w.r.t margin $m$ on both the base and novel classes of mini-ImageNet, respectively. 
As shown in Fig.~\ref{fig:discriminative_function}, for base classes (red curves), as the margin increases, the inter-class variance increases a lot, meanwhile the intra-class variance does not change much, so the features of base classes become more discriminative. This is widely observed in previous works \cite{deng2019arcface,wang2018cosface,liu2016largesoftmax}, and motivates them to introduce large and also positive margin to softmax loss for close-set scenarios.

But for novel classes (blue curves), the situation is just on the contrary. As the margin increases, the inter-class variance does not change much, but the intra-class variance increases a lot, so the features of base classes become less discriminative.
This indicates that larger margin may hurt the classification on the novel classes.
This is also verified in the real few-shot classification task, shown as red curves in Fig.~\ref{fig:teaser}, larger and positive margin will achieve worse performance of few-shot classification on novel classes.
Instead, the appropriate negative margin could achieve the best performance, which may lead to a better tradeoff on discriminability and transferability for novel classes.

\subsection{Intuitive Explanation}
\begin{figure*}[t]
    \centering
    \includegraphics[width=0.7\linewidth]{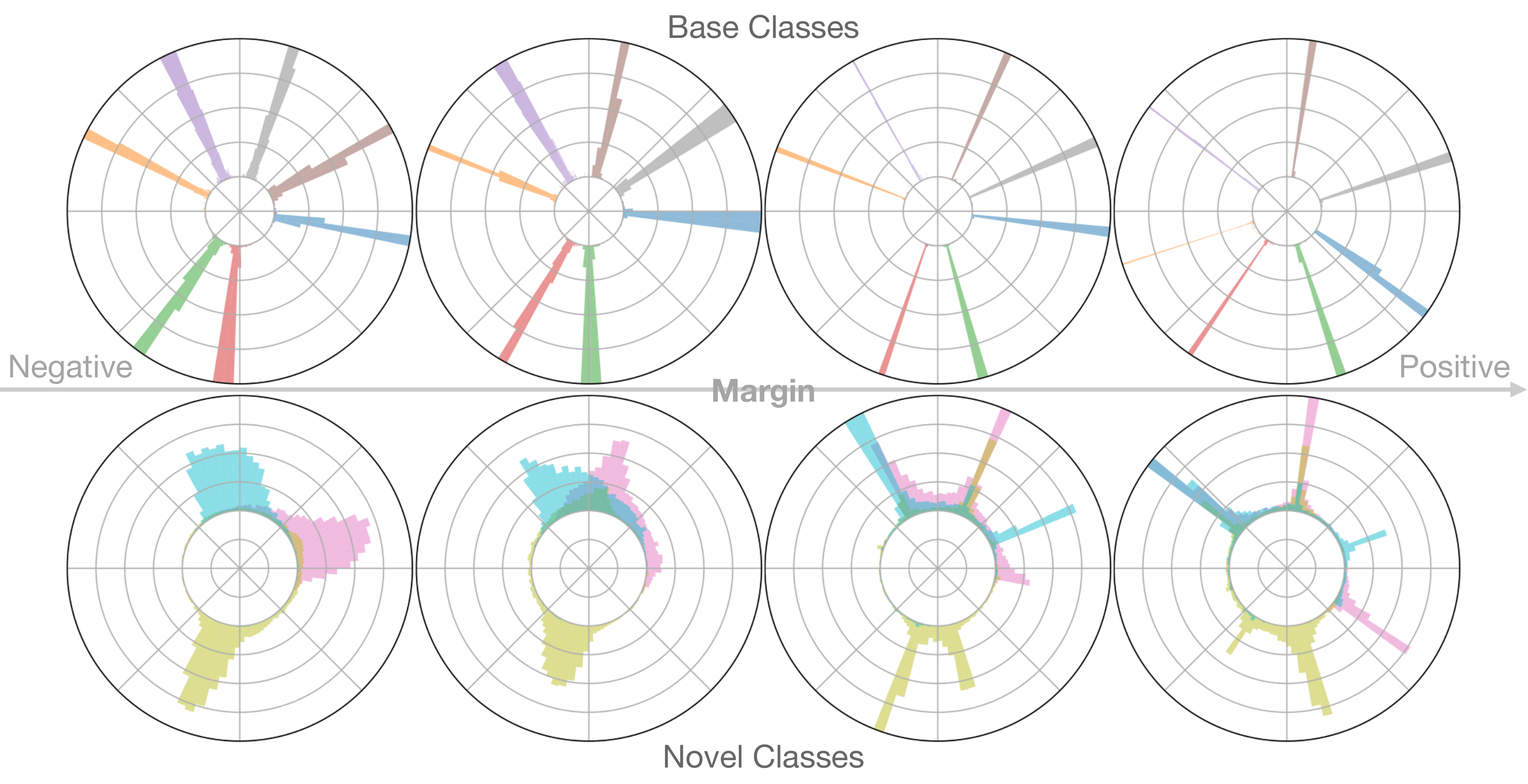}
    \vspace{-13pt}
    \caption{The visualizations of the data distributions on angular space with different margins, on base classes (the first row) or novel classes (the second row) of MNIST. Plots from left to right denotes the margins from negative to positive. For each figure, we plot the histogram of the occurrence for each angle. Different colors denote the data points belonging to different classes.}
    \label{fig:mnist}
    \vspace{-15pt}
\end{figure*}

To better understand how the margin works, we perform the visualization on the data distributions in the angular space trained on MNIST\footnote{This technique is widely used to characterize the feature embedding under the softmax-related objectives \cite{wang2018cosface,liu2017sphereface,zheng2018ring}.}, as shown in Fig.~\ref{fig:mnist}.
We choose seven classes as the base classes for pre-training, and adopt the other three classes as the novel classes.
We first train this deep model with 2-dimensional output features using cosine softmax loss with different margins on the base classes.
Then we normalize the 2-D features to obtain the direction of each data point, and visualize the count of each direction (also known as the data distributions in angular space) on both base (first row) and novel classes (second row) using the models trained with different margins.

As shown in the first row in Fig.~\ref{fig:mnist}, with larger and even positive margin (from left to right), the clusters for each training class are getting thinner and higher, and the angle differences between different class centers are getting larger.
This matches our previous observation in Fig.~\ref{fig:discriminative_function}, that enlarging the margin leads to the smaller intra-class variance and larger inter-class variance on the base classes. 

However, with larger margin, less data points would lie in the space far from all centers, which to some extent makes the output space much narrower.
As shown on the right side of the second row in Fig.~\ref{fig:mnist}, as novel classes are different to base classes, model with large margin may map the data points of the same class in novel classes to multiple peaks or clusters belonging to different base classes.
Then the intra-class variance for novel classes would increase accordingly, making the classification of novel classes more difficult. 
Instead, as shown on the left side of second row in Fig.~\ref{fig:mnist}, the appropriate negative margin would not enforce the data points in novel classes too close to the training center, and may alleviate the multi-peak issue, which could benefit the classification on novel classes. 

\subsection{Theoretical Analysis}
After giving the intuitive explanation that why negative margin works well on novel classes, we then prove this claim theoretically.
Denote the parameter of the classifier joint pre-trained with backbone on base classes with margin $m$ as ${\bf W}(m)$, the probability of a sample in the novel category $j$ classified by pre-trained backbone $f_{\theta(m)}$ and classifier $W(m)$ as a base category $k$ is
\begin{equation}
{\footnotesize
    P_{jk}(m) = \frac{1}{|I_{j}|} \sum_{(x_i, y_i) \in I_{j}} \frac{\exp \left(\beta \mathtt{s}(f_{\theta(m)}(x), W_{k}(m)) \right)}{\sum_{k^{\prime}=1}^{C^b} \exp \left( \beta \mathtt{s}(f_{\theta(m)}(x), W_{k^{\prime}}(m)) \right)},}
\end{equation}
where $\mathtt{s}(\cdot, \cdot)$ denotes the similarity function. The probability of a pair of samples in the same novel category $j$ classified into the same base class is {\small $P_j^s(m) = \sum_{k=1}^{C^b} P^2_{jk}(m)$}. And the average probability of {\small $P_j^s(m)$ is $P^s(m) = \frac{1}{|C^n|} \sum_{j=1}^{C^n} P_j^s(m)$}.

\begin{figure*}[tpb]
    \centering
    \subfigure[]{
        \includegraphics[width=0.52\textwidth]{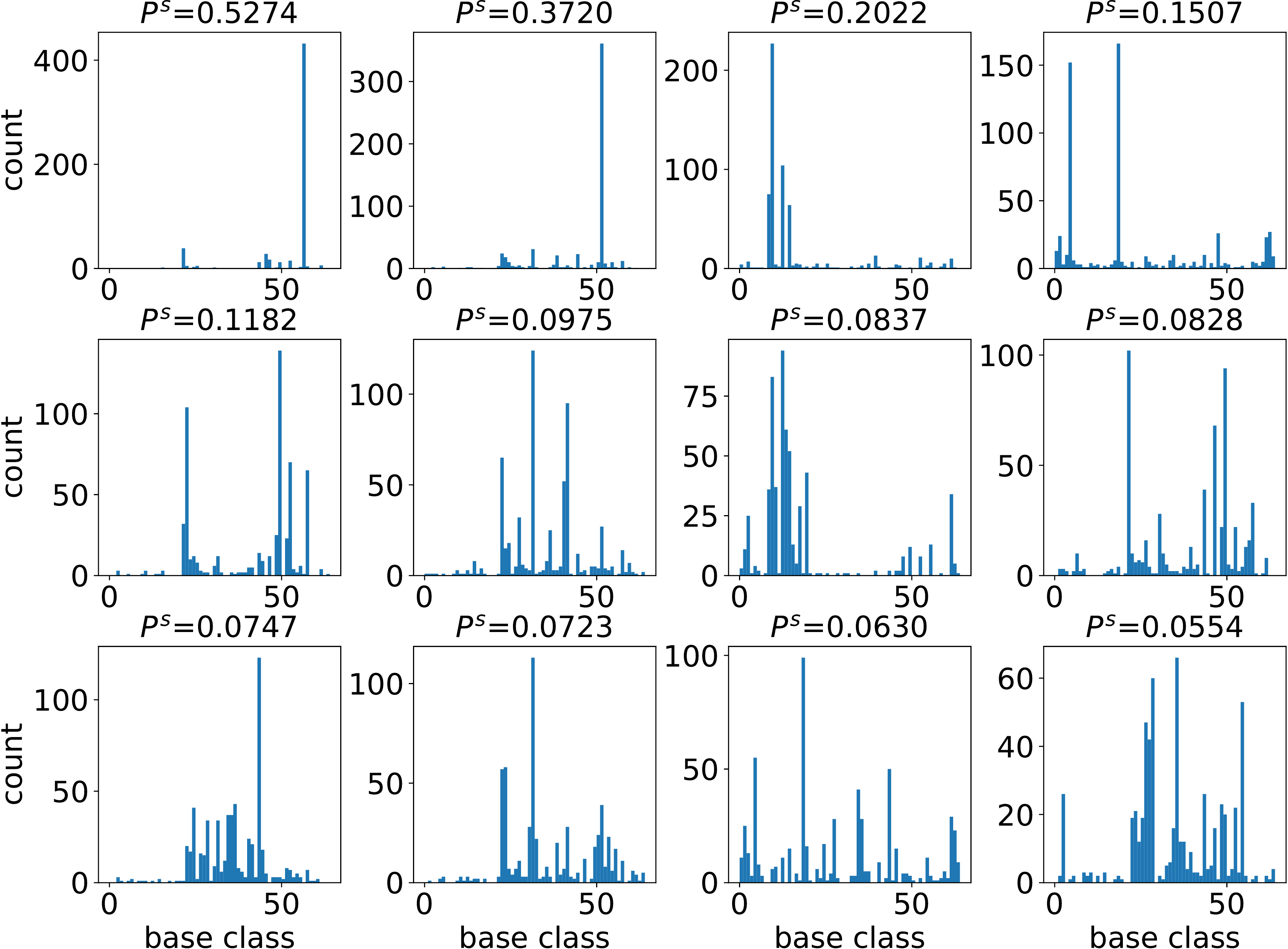}
    }
    \subfigure[]{
        \includegraphics[width=0.38\textwidth]{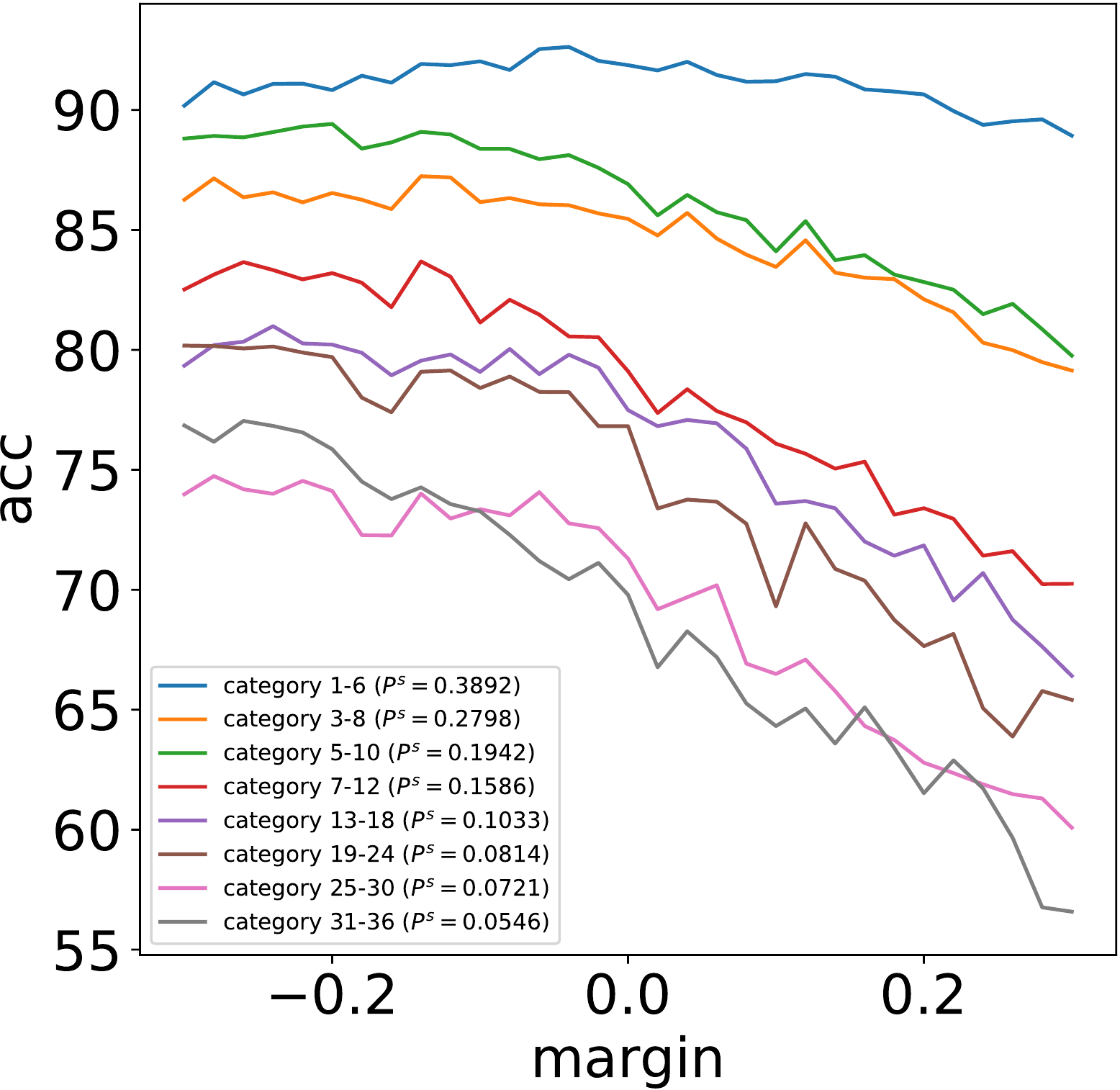}
    }
    \vspace{-15pt}
    \caption{We first sort the 36 novel classes according to the probability of sample pairs in the same novel class $j$ classified into the same base class $P_j^s$ (one of every 3 categories are plotted for clarity) on mini-ImageNet. For each novel class, (a) shows the histogram of samples in this class to be classified to 64 base classes.
    (b) shows the accuracy curves w.r.t different margins for novel classes with different averaged $P^s$.}
    \label{fig:assumption}
    \vspace{-15pt}
\end{figure*}

\noindent \textbf{Proposition}. Assuming discriminative function for the base classes $\phi(I^b, m)$ is a monotonic increasing function w.r.t margin parameter $m$, and then we denote $\phi^{-1}(I^b, m_1) - \phi^{-1}(I^b, m_2) = r \cdot (m_2 - m_1)$, where $m_2 > m_1$ and $r > 0$ is a scale variable. $\psi(m) = D_{\text{inter}}(I^n, m)/D_{\text{inter}}(I^b, m)$ is a monotonic decreasing function and we denote $\psi(m_1) - \psi(m_2) = t \cdot (m_2 - m_1), t > 0$. $\forall 0< P^s < \frac{t}{t(1-\phi^{-1}(I^b, m_1))+r\psi(m_1)}$, we have $\phi(I^n, m_2) < \phi(I^n, m_1)$.

\noindent \textbf{Proof}.
Since $D_{\text{intra}}(I^n, m) = P^s D_{\text{intra}}(I^b, m) + (1-P^s) D_{\text{inter}}(I^b, m) $, substituting it into the $ \phi(I^n, m_2) = \frac{D_{\text{inter}}(I^n, m_2)}{D_{\text{intra}}(I^n, m_2)} <  \frac{D_{\text{inter}}(I^n, m_1)}{D_{\text{intra}}(I^n, m_1)} = \phi(I^n, m_1)$, we have

\begin{equation}
    \footnotesize
    \begin{aligned}
        & \phi(I^n, m_2) < \phi(I^n, m_1) \\
        \Leftrightarrow & \frac{D_{\text{inter}}(I^n, m_2)}{P^s D_{\text{intra}}(I^b, m_2) + (1-P^s) D_{\text{inter}}(I^b, m_2)} <  \frac{D_{\text{inter}}(I^n, m_1)}{P^s D_{\text{intra}}(I^b, m_1) + (1-P^s) D_{\text{inter}}(I^b, m_1)} \\
        \Leftrightarrow &  P^s < 
            \frac{
                \frac{D_{\text{inter}}(I^n, m_1)}
                     {D_{\text{inter}}(I^b, m_1)} - 
                \frac{D_{\text{inter}}(I^n, m_2)}
                     {D_{\text{inter}}(I^b, m_2)} 
            }{ 
                \frac{D_{\text{inter}}(I^n, m_1)}
                     {D_{\text{inter}}(I^b, m_1)} 
                \cdot \left(1 - 
                    \frac{D_{\text{intra}}(I^b, m_2)}
                         {D_{\text{inter}}(I^b, m_2)} 
                \right)
                -
                \frac{D_{\text{inter}}(I^n, m_2)}
                     {D_{\text{inter}}(I^b, m_2)} 
                \cdot \left(1 - 
                    \frac{D_{\text{intra}}(I^b, m_1)}
                         {D_{\text{inter}}(I^b, m_1)} 
                \right) 
            } \\
        \Leftrightarrow &  P^s < \frac{ \psi(m_1) - \psi(m_2) }{ \psi(m_1) (1 - \phi^{-1}(I^b, m_2)) - \psi(m_2) (1 - \phi^{-1}(I^b, m_1)))} \\
        \Leftrightarrow &  P^s < \frac{t}{t(1-\phi^{-1}(I^b, m_1))+r\psi(m_1)}
    \end{aligned}
    \normalsize
\end{equation}

The above proposition proves that the discriminative function on the novel classes $\phi(I^n, m)$ is a monotonic decreasing function w.r.t $m$ under proper assumption and a measurable condition about the similarity between base and novel classes using $P^s$. The proposition indicates that an appropriate value of ``negative'' margin could work well for discriminating the samples in novel classes.

Fig.~\ref{fig:assumption} shows the actual behavior of mini-ImageNet dataset. We first sort the 36 novel classes according to the probability of sample pairs in the same novel class $j$ classified into the same base class $P_j^s$ (one of every 3 categories are plotted for clarity) on mini-ImageNet. And the histograms of the samples in novel classes to be classified to 64 base classes is shown in Fig.~\ref{fig:assumption}(a). Fig.~\ref{fig:assumption}(b) shows the accuracy curves w.r.t different margins for novel classes with different averaged $P^s$. With smaller $P^s$, the histograms of novel classes become more diverse (shown in Fig.~\ref{fig:assumption}(a)) and their accuracies become lower (shown in Fig.~\ref{fig:assumption}(b)). Importantly, most subsets of novel classes favor negative margins, implying the condition in the Proposition is not hard to reach.

\begin{figure*}[t]
    \centering
    \includegraphics[width=1.0\linewidth]{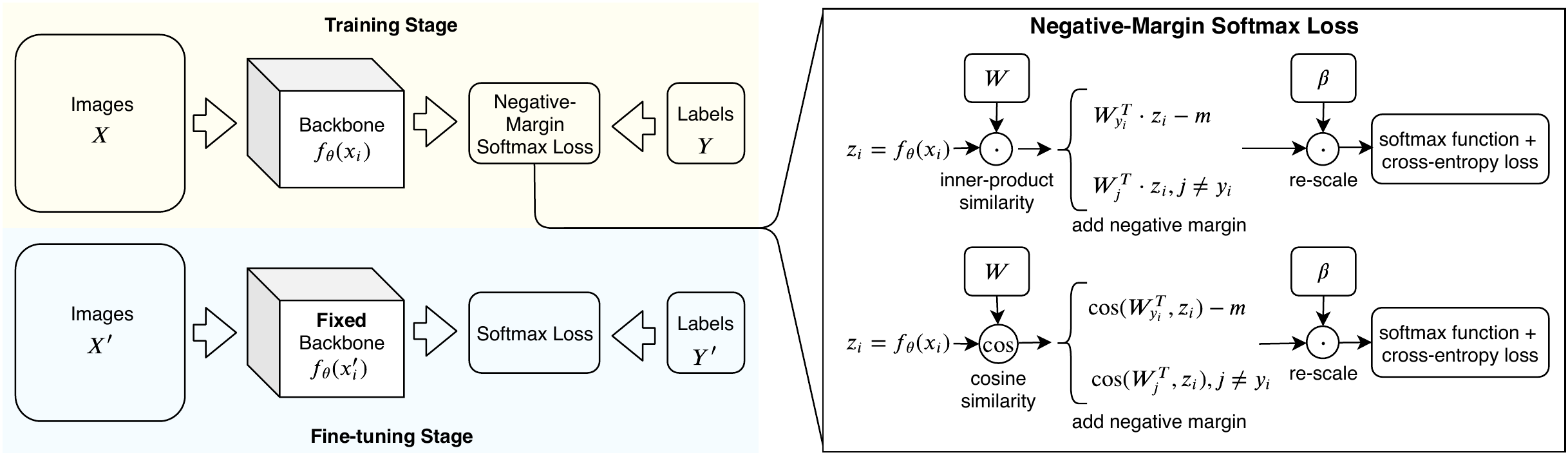}
    \vspace{-20pt}
    \caption{Overview of our proposed approach, which consists of two stages, pre-training (learning metrics with sufficient annotated examples on training classes) and fine-tuning (learning a classifier on novel classes with few labeled examples). The negative-margin softmax loss is integrated in pre-training for learning more transferable features for novel classes, with two types of similarities, inner-product and cosine similarity.}
    \label{fig:arch}
\end{figure*}

\subsection{Framework}\label{sec:framework}
Following the standard transfer learning paradigm \cite{yosinski2014howtransferable,donahue2014decaf}, we adopt a two-stage training pipeline for few-shot classification, as shown in Figure~\ref{fig:arch}, including pre-training stage to perform metric learning on the abundant labeled data in base classes, and fine-tuning stage to learn a classifier to recognize novel classes.
This pipeline is widely adopted in recent few-shot learning methods \cite{chen2019closerfewshot,dhillon2019baseline,mangla2019charting}.

In the pre-training stage, we aim at training the backbone network $f_{\theta}(\cdot)$ with abundant labeled data $I^b$ in base classes, driven by metric learning loss, such as softmax loss in \cite{chen2019closerfewshot}.
In our paper, we adopt the negative-margin softmax loss, which could learn more transferable representations for few-shot learning.
In the fine-tuning stage, as there are only few labeled samples in $I^n$ for training (e.g. 5-way 1-shot learning only contains 5 training samples), we follow \cite{chen2019closerfewshot} to fix the parameters of the backbone $f_{\theta}(\cdot)$, and only train a new classifier from scratch by the softmax loss. 
Note that, the computation of similarity (such as inner-product similarity or cosine similarity) in softmax loss is the same as that in the pre-training stage.

\section{Experiments}

\subsection{Setup}
\textbf{Datasets and scenarios.}
Following \cite{chen2019closerfewshot}, we address the few-shot classification problem under three different scenarios: \textbf{(1)} generic object recognition; \textbf{(2)} fine-grained image classification; and \textbf{(3)} cross-domain adaptation.

For the generic scenario, the widely-used few-shot classification benchmark: mini-ImageNet, is used to evaluate the effectiveness of the proposed Negative-Margin Softmax Loss. The mini-ImageNet dataset, firstly proposed by \cite{vinyals2016matching}, consists of a subset of 100 classes from the ILSVRC-2012 \cite{deng2009imagenet}, and contains 600 images for each classes. Following the commonly-used evaluation protocol of \cite{ravi2016optimization}, we split the 100 classes into 64 base, 16 validation, and 20 novel classes for pre-training, validation, and testing.

For the fine-grained image classification, we use CUB-200-2011 dataset \cite{WahCUB_200_2011} (hereinafter referred as CUB), which consists of 200 classes and 11,788 images in total. Following the standard setting of \cite{hilliard2018few}, we split the classes in the dataset into 100 base classes, 50 validation classes, and 50 novel classes.

For the cross-domain adaptation scenario, we use mini-ImageNet $\rightarrow$ CUB \cite{chen2019closerfewshot}, in which the 100 classes in mini-ImageNet, the 50 validation and 50 novel classes in CUB are adopted as base, validation and novel classes respectively, to evaluate the performance of the proposed Negative-Margin Softmax Loss in the presence of domain shift.

\noindent \textbf{Implementation details.}
For fair comparison, we evaluate our model with four commonly used backbone networks, namely Conv-4~\cite{vinyals2016matching}, ResNet-12~\cite{oreshkin2018tadam}, ResNet-18~\cite{chen2019closerfewshot} and WRN-28-10~\cite{mangla2019charting,Zagoruyko2016WRN}. Besides the differences in network depth and architecture, the expected input size of Conv-4 and ResNet-12 is 84$\times$84, and that of ResNet-18 is 224$\times$224, while WRN-28-10 takes 80$\times$80 images as input.

Our implementation is based on PyTorch~\cite{paszke2017automatic}. In the training stage, the backbone network and classifier are trained from scratch, with a batch size of 256. The models are trained for 200, 400 and 400 epochs in the CUB, mini-ImageNet and mini-ImageNet $\rightarrow$ CUB, respectively. 
We adopt the Adam~\cite{kingma2014adam} optimizer with initial learning rate 3e-3 and cosine learning rate decay~\cite{loshchilov2016sgdr}.
We apply the same data argumentation as \cite{chen2019closerfewshot}, including random cropping, horizontal flipping and color jittering.

In the fine-tuning stage, each episode contains 5 classes and each class contains 1 or 5 support images to train a new classifier from scratch and 16 query images to test the accuracy. The final performance is reported as the mean classification accuracy over 600 random sampled episodes with the 95\% confidence interval.
Note that all the hyper-parameters are determined by the performance on the validation classes.

\subsection{Results}
\textbf{Results on mini-ImageNet.}
For the generic object recognition scenario, we evaluate our methods on the widely-used mini-ImageNet dataset. 
For fair comparison with existing methods which uses different network architecture as backbone, we evaluate our methods with all four commonly used backbone networks.
The 5-way 1-shot and 5-shot classification results on the novel classes of the mini-ImageNet dataset are listed in Table~\ref{tab:mini-imagenet}.
We find that by simply adopting appropriate negative margin in standard softmax loss, our Neg-Softmax achieves competitive results with the existing state-of-the-art methods.
It is worth noting that our Neg-Cosine achieves the state-of-the-art performance for both 1-shot and 5-shot settings on almost all four backbones on mini-ImageNet.

\begin{table*}[tbp]
\centering
\addtolength{\tabcolsep}{6.5pt}
\caption{Few-shot classification results on the mini-ImageNet dataset. $^\dagger$ indicates the method using the combination of base and validation classes to train the meta-learner}
\label{tab:mini-imagenet}
\footnotesize
\begin{tabular}{l|lcc}
\Xhline{2\arrayrulewidth}
\textbf{Backbone} & \textbf{Method}                & \textbf{1 shot}  & \textbf{5 shot}  \\ 
\hline 
\multirow{7}{*}{Conv-4}    
& MAML \cite{finn2017MAML}               &  48.70 $\pm$ 1.84 & 63.11 $\pm$ 0.92 \\
& ProtoNet \cite{snell2017prototypical}  &  49.42 $\pm$ 0.78 & 68.20 $\pm$ 0.66 \\
& MatchingNet \cite{vinyals2016matching} &  48.14 $\pm$ 0.78 & 63.48 $\pm$ 0.66 \\
& RelationNet \cite{sung2018relation}    &  50.44 $\pm$ 0.82 & 65.32 $\pm$ 0.70 \\
& MAML+Meta-dropout \cite{Lee2020Meta}    &  51.93 $\pm$ 0.67 & 67.42 $\pm$ 0.52 \\
& R2D2 \cite{bertinetto2018R2D2}         &  51.20 $\pm$ 0.60 & 68.80 $\pm$ 0.10 \\
\cline{2-4}
& Neg-Softmax (ours) & 47.65 $\pm$ 0.78 &  67.27 $\pm$ 0.66 \\
& Neg-Cosine (ours)  & \textbf{52.84 $\pm$ 0.76} & \textbf{70.41 $\pm$ 0.66} \\ 
\hline \hline
\multirow{5}{*}{ResNet-12} 
& SNAIL \cite{mishra2017simple}         & 55.71 $\pm$ 0.99 & 68.88 $\pm$ 0.92 \\
& TADAM \cite{oreshkin2018tadam}        & 58.50 $\pm$ 0.30 & 76.70 $\pm$ 0.30 \\
& MetaOptNet-SVM \cite{lee2019metaopt}  & 62.64 $\pm$ 0.61 & 78.63 $\pm$ 0.46 \\
\cline{2-4}
& Neg-Softmax (ours)    & 62.58 $\pm$ 0.82 & 80.43 $\pm$ 0.56 \\
& Neg-Cosine (ours)     & \textbf{63.85 $\pm$ 0.81} & \textbf{81.57 $\pm$ 0.56} \\ 
\hline \hline
\multirow{5}{*}{ResNet-18} 
& SNCA \cite{wu2018SNCA}                  & 57.80 $\pm$ 0.80 & 72.80 $\pm$ 0.70 \\
& Baseline \cite{chen2019closerfewshot}   & 51.75 $\pm$ 0.80 & 74.27 $\pm$ 0.63 \\
& Baseline++ \cite{chen2019closerfewshot} & 51.87 $\pm$ 0.77 & 75.68 $\pm$ 0.63 \\
\cline{2-4}
& Neg-Softmax (ours)    & 59.02 $\pm$ 0.81 & 78.80 $\pm$ 0.61 \\
& Neg-Cosine (ours)     & \textbf{62.33 $\pm$ 0.82} & \textbf{80.94 $\pm$ 0.59} \\ 
\hline \hline
\multirow{6}{*}{WRN-28-10} 
& Activation to Parameter$^\dagger$ \cite{qiao2018few} & 59.60 $\pm$ 0.41 & 73.74 $\pm$ 0.19 \\ 
& LEO$^\dagger$  \cite{rusu2018LEO}                    & 61.76 $\pm$ 0.08 & 77.59 $\pm$ 0.12 \\                
& Fine-tuning \cite{dhillon2019baseline}                & 57.73 $\pm$ 0.62 & 78.17 $\pm$ 0.49 \\
& Cosine + rotation \cite{gidaris2019boosting}          & \textbf{62.93 $\pm$ 0.45} & 79.87 $\pm$ 0.33 \\
\cline{2-4}
& Neg-Softmax (ours) & 60.04 $\pm$ 0.79 & 80.90 $\pm$ 0.60 \\
& Neg-Cosine (ours)  & 61.72 $\pm$ 0.81 & \textbf{81.79 $\pm$ 0.55} \\ 
\Xhline{2\arrayrulewidth}
\end{tabular}
\normalsize
\vspace{-5pt}
\end{table*}

\noindent \textbf{Results on CUB.}
On the fine-grained dataset CUB, we compared the proposed method with several state-of-the-art methods with ResNet-18 as backbone. The results are showed in Table~\ref{tab:CUB-cross}, in which the results of the comparison methods are directly borrowed from \cite{chen2019closerfewshot}.
It shows that the proposed Neg-Cosine outperforms all the comparison methods on both 1-shot and 5-shot settings.
Furthermore, Neg-Softmax also achieves highly competitive performance on both 1-shot and 5-shot settings.

\noindent \textbf{Results on mini-ImageNet $\rightarrow$ CUB.}
In the real-world applications, there may be a signification domain shift between the base and novel classes. So we evaluate our methods on a cross domain scenario: mini-ImageNet $\rightarrow$ CUB, where we pre-train the backbone on a generic object recognition dataset, and transfer it to a fine-grained dataset. We follow \cite{chen2019closerfewshot} to report the 5-shot results with ResNet-18 backbone, as shown in Table~\ref{tab:CUB-cross}. We can observe that both Neg-Softmax and Neg-Cosine are significantly better than all the comparison methods. Specifically, Neg-Softmax outperforms Baseline~\cite{chen2019closerfewshot}, the state-of-the-art method on the mini-ImageNet $\rightarrow$ CUB, by a large margin of $3.73\%$.

\subsection{Analysis}

This section presents a comprehensive analysis of the proposed approach. In the following experiments, we use Neg-Cosine with ResNet-18 backbone as default.

\noindent \textbf{Effects of negative margin.}
Table~\ref{tab:tab_neg_margin} shows the 1-shot and 5-shot accuracy of the standard softmax, cosine softmax and our proposed Neg-Softmax, Neg-Cosine on the validation classes of mini-ImageNet, CUB and mini-ImageNet $\rightarrow$ CUB.
By adopting appropriate negative margin, Neg-Softmax and Neg-Cosine yields significant performance gains over standard softmax loss and cosine softmax loss on all three benchmarks.
Interestingly, Neg-Cosine outperforms Neg-Softmax in the in-domain setting, such as mini-ImageNet and CUB, while Neg-Softmax could achieve better performance than Neg-Cosine in the cross-domain setting. This is also observed in \cite{chen2019closerfewshot}.


\begin{table}[tbp]
\centering
\addtolength{\tabcolsep}{0.5pt}
\caption{The few-shot classification accuracy on the novel classes (also known as test classes) of the CUB dataset and cross-domain setting with ResNet-18 as the backbone}
\label{tab:CUB-cross}
\footnotesize
\begin{tabular}{lccc}
\Xhline{2\arrayrulewidth}
\multirow{2}{*}{\textbf{Method}}    & \multicolumn{2}{c}{\textbf{CUB}} & \textbf{mini-ImageNet$\rightarrow$CUB}\\ 
 & \textbf{1 shot } & \textbf{5 shot } & \textbf{5 shot } \\ 
\hline
MAML \cite{finn2017MAML}               & 69.96 $\pm$ 1.01 & 82.70 $\pm$ 0.65 & 51.34 $\pm$ 0.72 \\
ProtoNet \cite{snell2017prototypical}  & 71.88 $\pm$ 0.91 & 87.42 $\pm$ 0.48 & 62.02 $\pm$ 0.70 \\
MatchingNet \cite{vinyals2016matching} & {72.36 $\pm$ 0.90} & 83.64 $\pm$ 0.60 & 53.07 $\pm$ 0.74 \\
RelationNet \cite{sung2018relation}    & 67.59 $\pm$ 1.02 & 82.75 $\pm$ 0.58  & 57.71 $\pm$ 0.73 \\
Baseline  \cite{chen2019closerfewshot}    & 65.51 $\pm$ 0.87 & 82.85 $\pm$ 0.55  & 65.57 $\pm$ 0.70 \\
Baseline++  \cite{chen2019closerfewshot}  & 67.02 $\pm$ 0.90 & 83.58 $\pm$ 0.54 & 62.04 $\pm$ 0.76  \\
\hline
Neg-Softmax (ours) & 71.48 $\pm$ 0.83 & 87.30 $\pm$ 0.48  & \textbf{69.30 $\pm$ 0.73}\\
Neg-Cosine (ours)  & \textbf{72.66 $\pm$ 0.85} & \textbf{89.40 $\pm$ 0.43}  & 67.03 $\pm$ 0.76 \\ 
\Xhline{2\arrayrulewidth}
\end{tabular}
\normalsize
\vspace{-5pt}
\end{table}

\begin{table*}[tb]	
\addtolength{\tabcolsep}{0.8pt}
\caption{The few-shot accuracy of standard softmax, cosine softmax and our proposed Neg-Softmax, Neg-Cosine on validation classes of three standard benchmarks}
\label{tab:tab_neg_margin}
\footnotesize
\scalebox{0.9}{
\begin{tabular}{c c cc c cc c cc}
\Xhline{2\arrayrulewidth}
\multirow{2.5}{*}{\textbf{Method}} & \phantom{a} & \multicolumn{2}{c}{\textbf{{\scriptsize mini-ImageNet}}} & \phantom{a} & \multicolumn{2}{c}{\textbf{{\scriptsize CUB}}} & \phantom{a} & \multicolumn{2}{c}{\textbf{{\scriptsize mini-ImageNet$\rightarrow$CUB}}} \\
\cmidrule{3-4} \cmidrule{6-7} \cmidrule{9-10}
 && \textbf{1 shot} & \textbf{5 shot} && \textbf{1 shot} & \textbf{5 shot} && \textbf{1 shot} & \textbf{5 shot} \\
\hline 
Softmax && 45.98$\pm$0.79 & 75.25$\pm$0.61 && 58.32$\pm$0.87 & 80.21$\pm$0.59 && 46.87$\pm$0.78 & 67.68$\pm$0.71 \\
Neg-Softmax && 56.95$\pm$0.82 & 78.87$\pm$0.57 && 59.54$\pm$0.88 & 80.60$\pm$0.57 && 47.74$\pm$0.73 & 68.58$\pm$0.70 \\
\hline 
Cosine && 59.49$\pm$0.90 & 79.58$\pm$0.59 &&66.39$\pm$0.93 & 82.17$\pm$0.58 && 42.96$\pm$0.76 & 61.99$\pm$0.75 \\
Neg-Cosine && 63.68$\pm$0.86 & 82.02$\pm$0.57 && 69.17$\pm$0.85 & 85.60$\pm$0.56 && 44.51$\pm$0.85 & 64.04$\pm$0.75 \\
\Xhline{2\arrayrulewidth}
\end{tabular}}
\normalsize
\vspace{-10pt}
\end{table*}

\begin{figure*}[!tbp]
    \centering
    \subfigure[Neg-Cosine]{
        \includegraphics[width=0.475\textwidth]{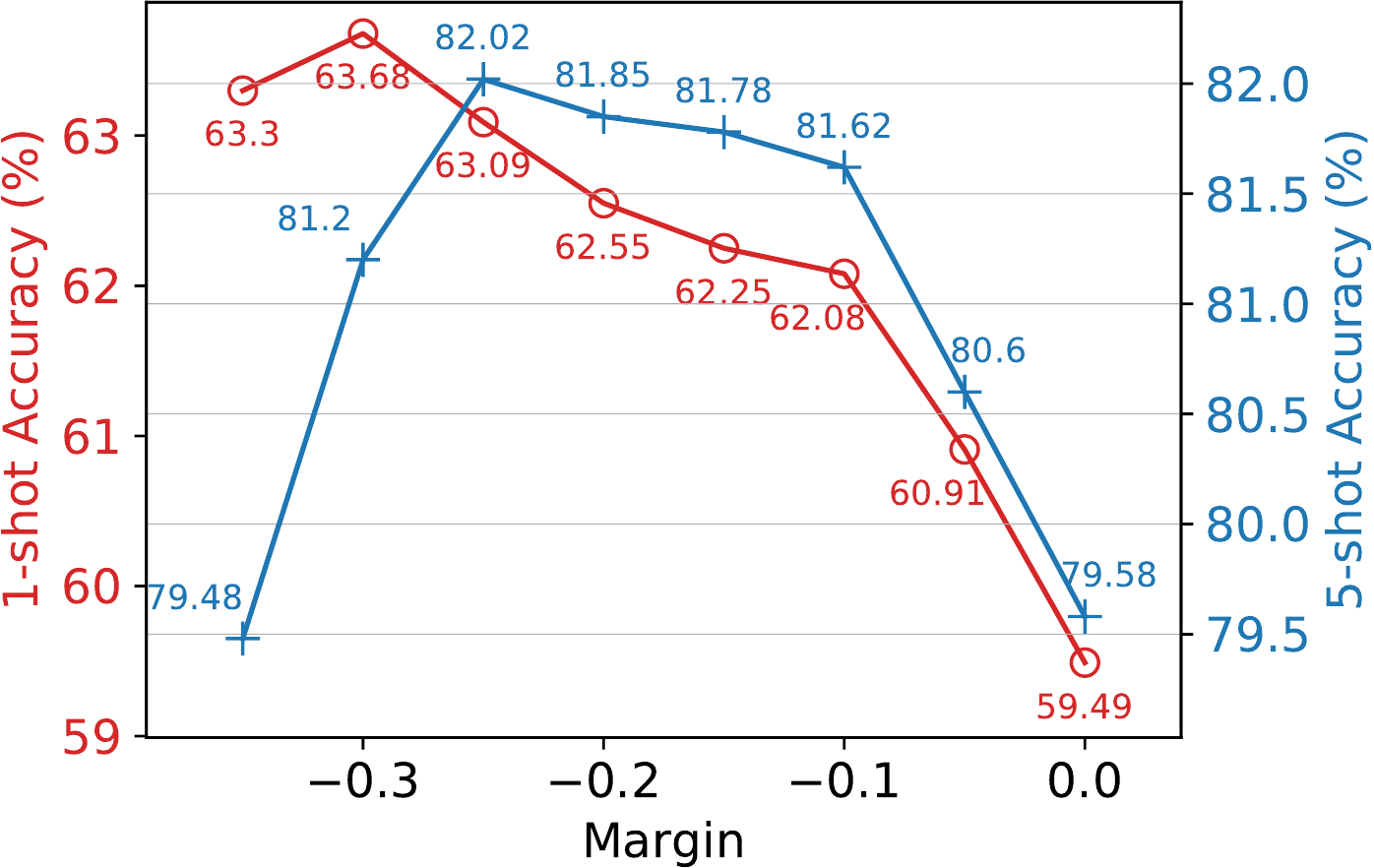}
    }
    \subfigure[Neg-Softmax]{
        \includegraphics[width=0.475\textwidth]{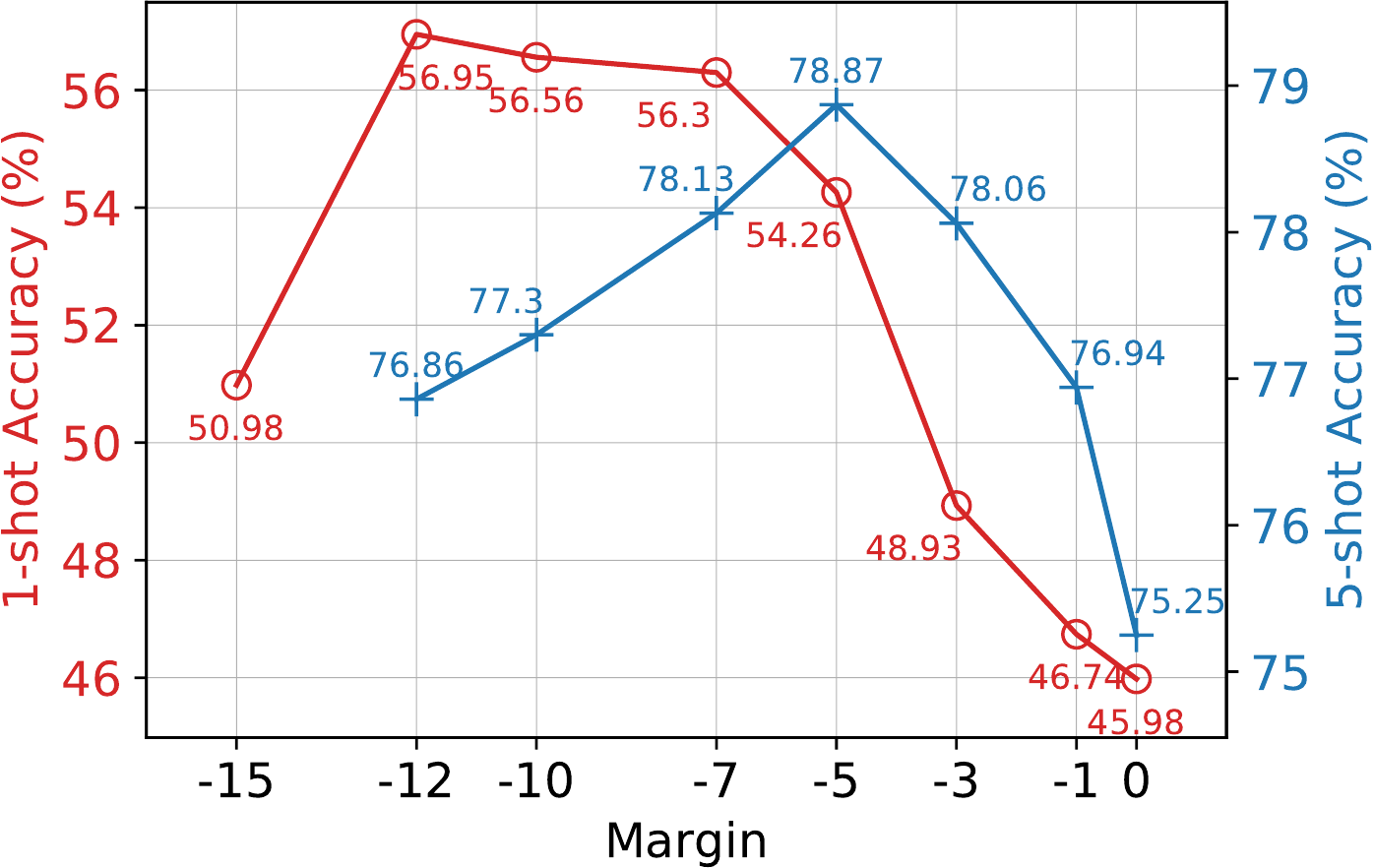}
    }
    \vspace{-10pt}
    \caption{The 1-shot (on red) and 5-shot (on blue) accuracy on validation classes of mini-ImageNet w.r.t different margins in Neg-Cosine and Neg-Softmax}
    \label{fig:margin}
\end{figure*}

\noindent \textbf{Accuracy w.r.t different margins.}
Figure~\ref{fig:margin} shows the 1-shot accuracy and 5-shot accuracy on validation classes of mini-ImageNet dataset w.r.t different margins in Neg-Cosine and Neg-Softmax. As we expect, as the margin gets negative and smaller, both the 1-shot accuracy and 5-shot accuracy of Neg-Cosine and Neg-Softmax first increase and then decrease, demonstrating a desirable bell-shaped curve.
Hence, adopting appropriate negative margin yields significant performance gains over both standard softmax loss and cosine softmax loss on 1-shot and 5-shot classification of mini-ImageNet.

\newcommand{\twoline}[2]{\begin{tabular}[c]{@{}l@{}}\textbf{#1} \\ \textbf{#2}\end{tabular}}
\begin{table}[t]
\centering
\addtolength{\tabcolsep}{6.5pt}
\caption{Test accuracy on 5-way mini-ImageNet of various regularization techniques}
\label{tab:ablation}
\footnotesize
\vspace{-5pt}
\begin{tabular}{lll|cc}
\Xhline{2\arrayrulewidth}
\twoline{negative}{margin} & \twoline{weight}{decay} & \twoline{drop}{block} & \textbf{1 shot} & \textbf{5 shot} \\
\hline 


&& & 54.51 $\pm$ 0.79 &	75.70 $\pm$ 0.62 \\
\checkmark & & & 60.25 $\pm$ 0.81 &	80.07 $\pm$ 0.58 \\
\checkmark & \checkmark & & 62.21 $\pm$ 0.83 &	80.81 $\pm$ 0.59 \\
\checkmark & \checkmark & \checkmark & 62.33 $\pm$ 0.82 & 80.94 $\pm$ 0.59 \\
\Xhline{2\arrayrulewidth}
\end{tabular}
\normalsize
\end{table}

%

\noindent \textbf{Various regularization techniques.}
Table~\ref{tab:ablation} shows the importance of regularizations on Neg-Cosine, which reveals that integrating various regularization techniques steadily improves the 1-shot and 5-shot test accuracy on mini-ImageNet benchmark. 
Firstly, by simply adopting negative margin, the test accuracy increased by 5.74\% and 4.37\% on the 1-shot and 5-shot settings, respectively.
Based on our approach, weight decay and DropBlock could further improve the performance.
After integrating all regularizations together, our method achieves state-of-the-art accuracy of $62.33\%$ and $80.94\%$ for the 1-shot and 5-shot settings respectively on novel classes of mini-ImageNet.

\begin{figure}[t]
    \centering
    \includegraphics[width=0.65\linewidth]{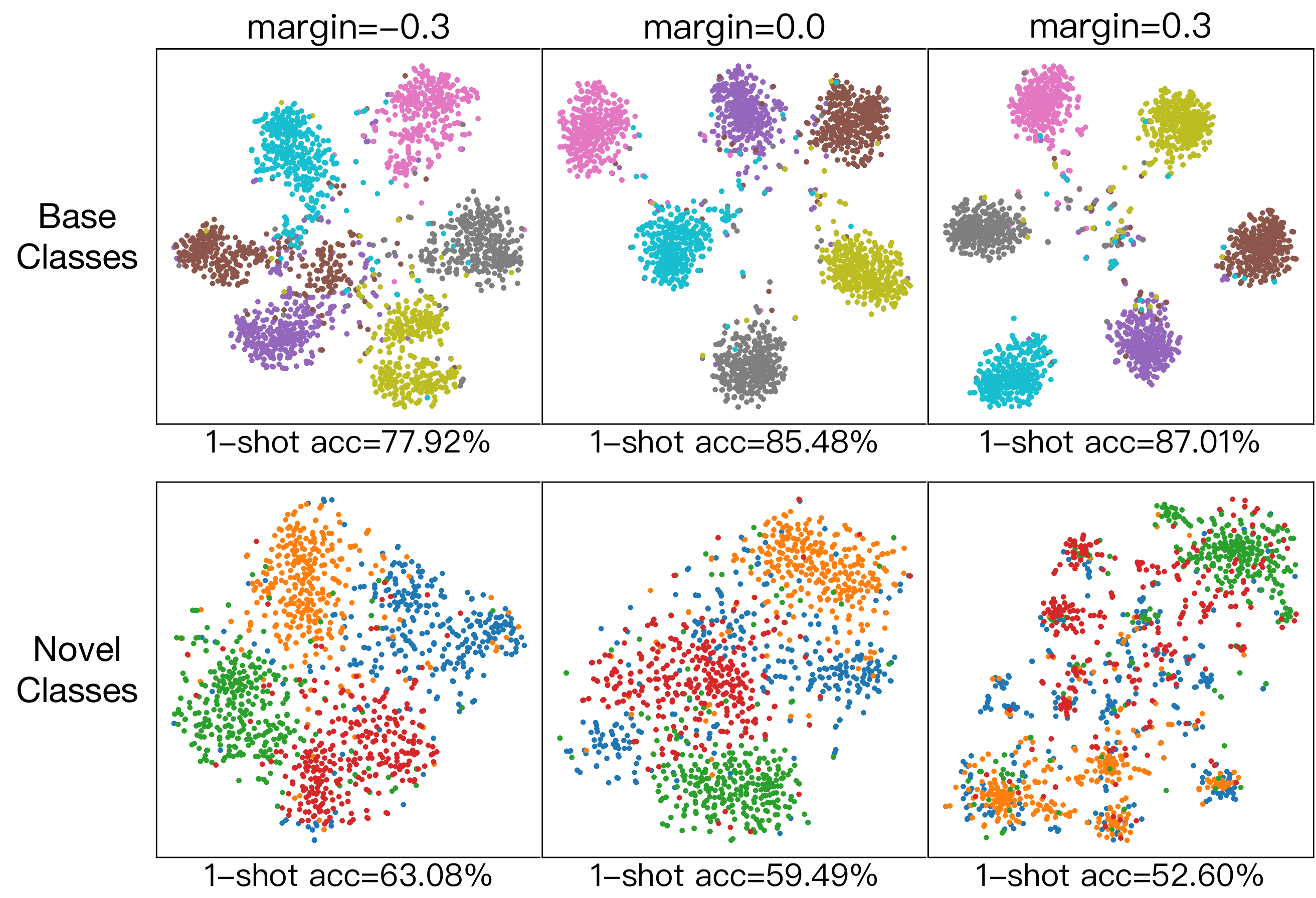}
    \caption{The t-SNE visualizations of the feature embeddings and the corresponding 1-shot accuracy in the base and novel classes of mini-ImageNet dataset for the softmax loss with negative, zero and positive margin respectively}
    \label{fig:tsne}
\vspace{-5pt}
\end{figure}

\noindent \textbf{T-SNE visualization.}
Fig.~\ref{fig:tsne} shows the t-SNE \cite{cite:tsne} visualizations of the feature embedding and the corresponding 1-shot accuracy in the base and novel classes of mini-ImageNet dataset for negative, zero and positive margin respectively.
As shown in the first row in Fig.~\ref{fig:tsne}, compared with negative margin, the feature embedding of zero and positive margin exhibit more discriminative structures and achieve better 1-shot accuracy on the base classes. 
However, the second row in Figure~\ref{fig:tsne} shows that enlarging the margin parameter would break the cluster structure of the novel classes and make the classification of novel classes harder.
Instead, the appropriate negative margin could tradeoff the discriminability and transferability of deep features in the novel classes, and retain the better cluster structure for novel classes. Thus the few-shot classification accuracy of negative margin is better than that of zero and positive margin.

\begin{figure*}
\centering
\includegraphics[width=0.6\linewidth]{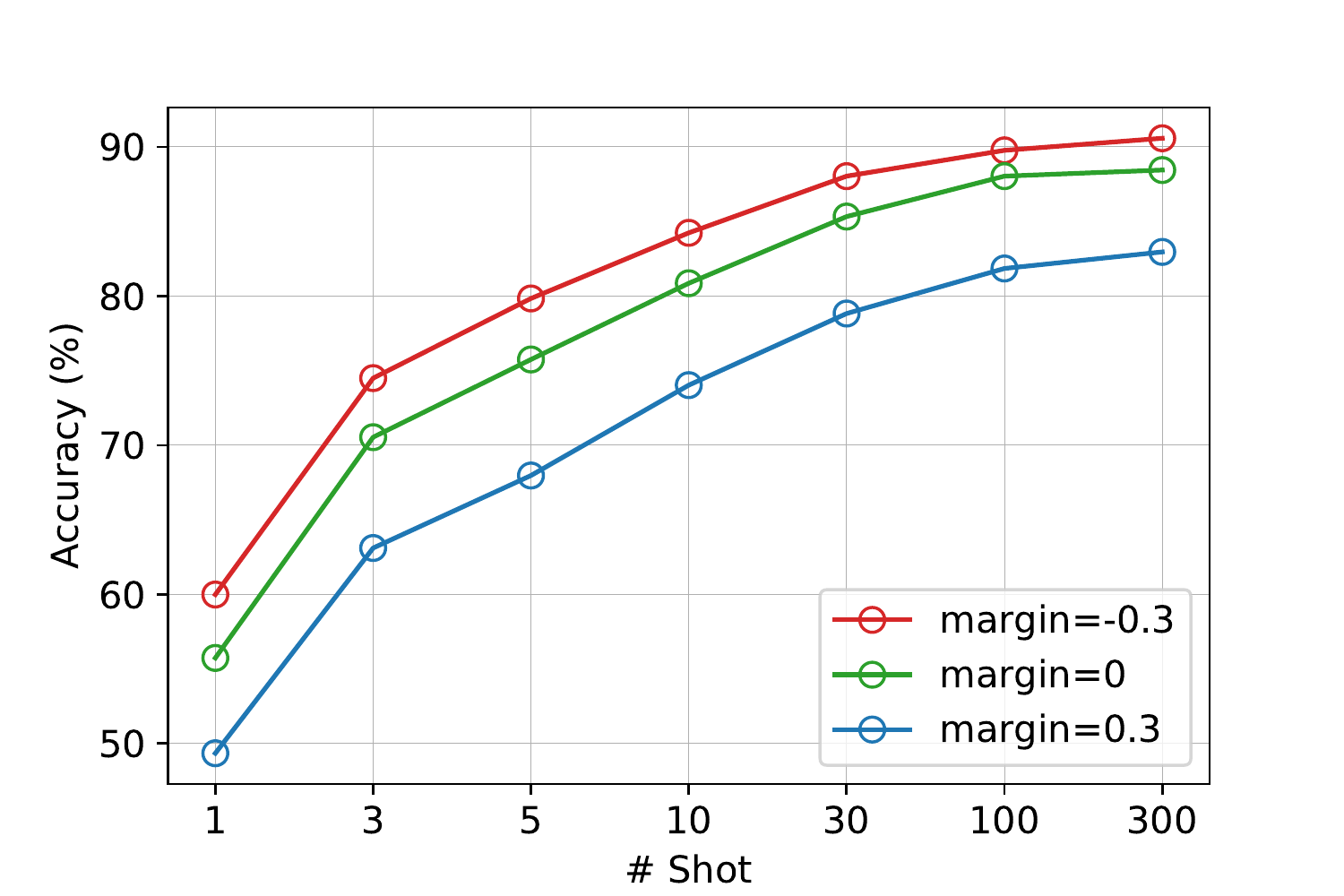}
    \vspace{-5pt}
    \caption{Accuracy w.r.t \# shots of validation classes on the mini-ImageNet dataset for margin = -0.3, 0 and 0.3}
    \label{fig:n_shot}
\end{figure*}

\noindent \textbf{More shots.}
We conduct an experiment by varying the number of shots from 1 (few shot) to 300 (many shot) and report the classification accuracy of the validation classes on the mini-ImageNet dataset in Figure~\ref{fig:n_shot}. It shows that the test accuracy of margin=$-0.3$ is consistently higher than that of margin=$0$ from 1-shot to 300-shot settings, which prove that the negative margin could benefit the open-set scenarios with more shots. 
But there is another trend we cannot ignore, the gap between margin=$-0.3$ and margin=$0$ is getting smaller when the number of shots increase.
In other words, when sufficient data is available in the novel classes, the influence of the metrics learned from the base classes to the novel classes is weakened.

\section{Conclusion}
In this paper, we unconventionally propose to adopt appropriate negative-margin to softmax loss for few-shot classification, which surprisingly works well for the open-set scenarios of few-shot classification.
We then provide the intuitive explanation and the theoretical proof to understand why negative margin works well for few-shot classification. 
This claim is also demonstrated via sufficient experiments. With the negative-margin softmax loss, our approach achieves the state-of-the-art performance on all three standard benchmarks of few-shot classification.
In the future, the negative margin may be applied in more general open-set scenarios that do not restrict the number of samples in novel classes.

\appendix
\section{Instantiations of Negative-Margin Loss}

Eqn.~\ref{eqn:nm_softmax_general} provides a general formulation for the margin softmax loss. Here, we provide detailed formulation of the two instantiations: negative-margin softmax loss and negative-margin cosine softmax loss as:

\begin{equation}
\small
L =  - \frac{1}{N}\sum\limits_{i = 1}^N {\log \frac{{{e^{\beta  \cdot \left( {W_{{y_i}}^T{z_i} - m} \right)}}}}{{{e^{\beta  \cdot \left( {W_{{y_i}}^T{z_i} - m} \right)}} + \sum\limits_{j = 1,j \ne {y_i}}^C {{e^{\beta  \cdot \left( {W_j^T{z_i}} \right)}}} }}},
\end{equation}
\begin{equation}
\small
    L =  - \frac{1}{N}\sum\limits_{i = 1}^N {\log \frac{{{e^{\beta  \cdot \left( {\cos \left( {{W_{{y_i}}},{z_i}} \right) - m} \right)}}}}{{{e^{\beta  \cdot \left( {\cos \left( {{W_{{y_i}}},{z_i}} \right) - m} \right)}} + \sum\limits_{j = 1,j \ne {y_i}}^C {{e^{\beta  \cdot \cos \left( {{W_j},{z_i}} \right)}}} }}} ,
\end{equation}
where $m\le0$ is the margin parameter. Note that the formulations are the same as the \emph{large}-margin softmax loss in \cite{liu2016largesoftmax} and the large-margin cosine loss in \cite{wang2018cosface}, but we restrict the margin $m$ as a non-positive value ($m\le0$) while the original formulations restrict the margin $m$ as a non-negative value ($m\ge 0$). 

We follow the pre-training and fine-tuning pipeline introduced in Sec.~\ref{sec:framework}, and the proposed negative-margin (cosine) softmax loss is applied in the \emph{pre-training} stage, as illustrated in Figure \ref{fig:arch}. Note we do not apply the negative-margin loss to the \emph{pre-training} stage, where we find regular (cosine) softmax loss (margin $m=0$) performs well, as detailed in the following section.

\section{The effects of margin in the fine-tuning stage}

\begin{figure}[!ht]
    \centering
    \includegraphics[width=0.5\textwidth]{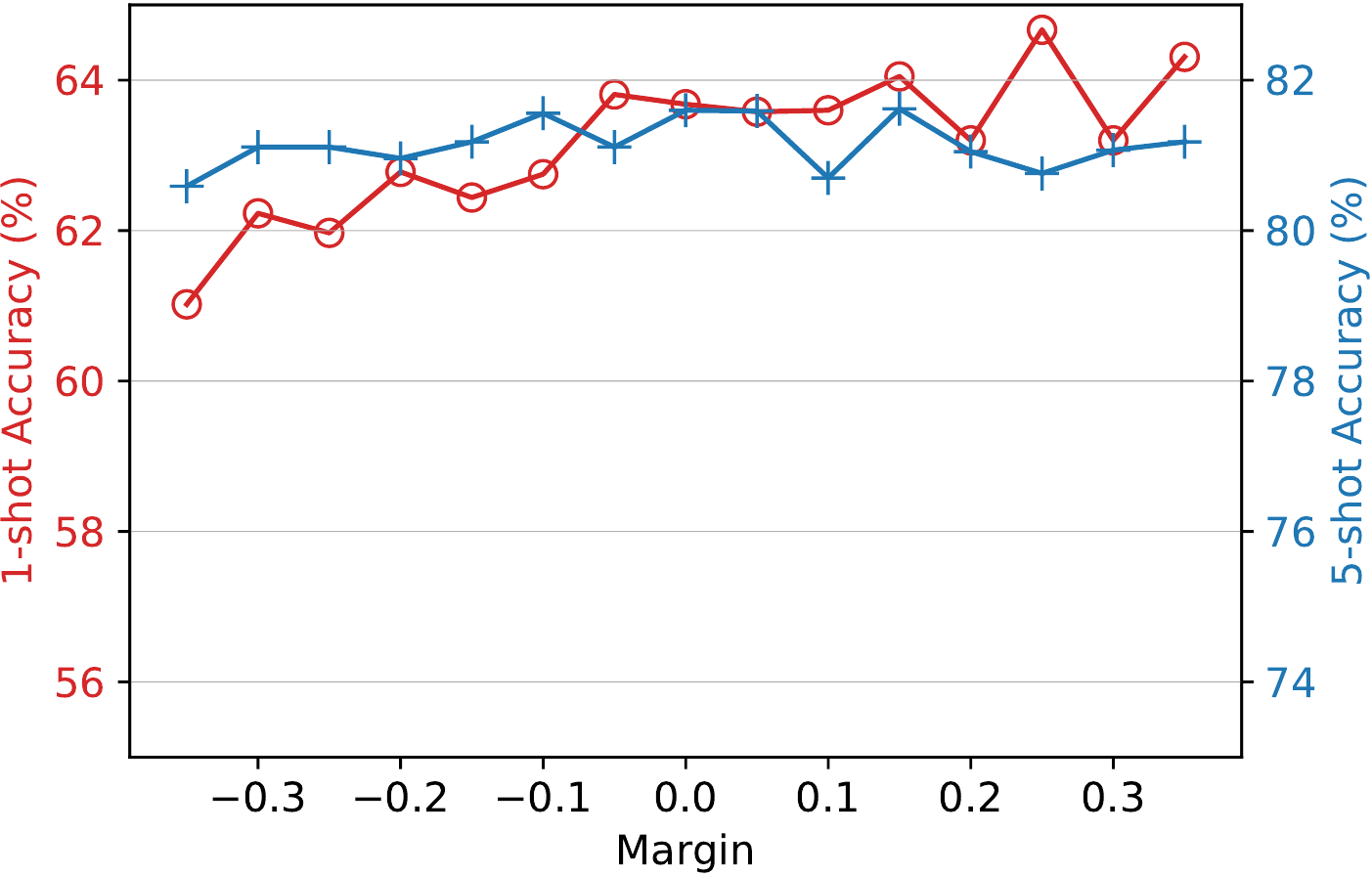}
    \caption{The one-shot  (in red) and five-shot (in blue) accuracy on validation classes w.r.t different margins in cosine softmax loss applied in fine-tunning stage on mini-ImageNet. The margin parameter in the pre-training stage is fixed to $-0.3$ and the backbone is ResNet-18.} 
    \label{fig:acc_finetune_margin}
\end{figure}

In the main part of the paper, we show that applying the negative-margin softmax loss into the pre-training stage leads to better discrimination of novel classes.
In this section, we investigate the effects of the margin parameter on the fine-tuning stage.
As shown in Fig.~\ref{fig:acc_finetune_margin}, the 5-shot classification accuracy in the validation classes is insensitive when varying margin values, while that of 1-shot classification increases marginally when increasing margin values. Such behavior is different from the effects of the margin parameter in the pre-training stage, indicating that the negative margin mainly effects in open-set scenarios. Also note the accuracy is much more insensitive w.r.t margin parameter than that of pre-training and fine-tuning both using base classes (the blue curves in Fig~\ref{fig:teaser}), probably because the feature is fixed in the fine-tuning stage and different margin values can all find good discrimination planes well.

\section{Relationship between negative margin and label smoothing} 

The same as negative margin loss, the label smoothing \cite{szegedy2016rethinking} technique also aims at facilitating the difficulty of equalising softmax outputs and the binary ground-truth labels. Technically, while the label smoothing technique changes the ground truth target labels from binary values to soft ones which alleviate the equalising of softmax outputs and the binary ground-truth labels, the negative margin loss modify the other side of softmax outputs. Although resulting in similarly smaller loss than regular training, they  perform very differently on training classes and novel classes.

In softmax based image classification, the final prediction is the class with the largest softmax probability and the accuracy is 100\% if all predictions match the ground truth label. The softmax output for the ground truth label is not necessarily as 1, but is correct as long as it is larger than the softmax outputs of other classes. In another word, the cross entropy loss which encourages the softmax output of the right class to be exactly 1 is stricter than the real target of doing right classification. The label smoothing technique relaxes the ground-truth labels as soft ones, which shrinks the gap between cross-entropy loss and the real classification target. As a result, it usually achieves higher accuracy on the validation images of training classes, indicating better discrimination of training classes. On the contrary, the negative margin technique enlarges the gap between loss and real classification accuracy. It usually results in lower accuracy on the validation images of training classes, indicating lower discrimination of training classes.

Table~\ref{tab:label_smoothing} investigate the effects of two techniques by applying few-shot fine-tuning on either validation images (ILSVRC-2012 images of the base classes but not in mini-ImageNet) of base classes or images of val classes.  For label smoothing, $\epsilon$ is set as 0.05. For negative margin, $m = -0.3$.
It can be seen that the label smoothing technique improves the performance on the base classes by 3.48\% and 0.21\% for the 1-shot and 5-shot setting respectively, but has 1.47\% and 3.03\% accuracy drop on the validation classes for the 1-shot and 5-shot settings, respectively.
On the contrary, the negative margin technique has lower performance on the validation images of training classes but benefits both the 1-shot and 5-shot classification accuracy on the validation classes.

These results are in accord with the previous analysis that the different effects of the negative margin technique and the label smoothing technique. While the
label smoothing technique tends to improve the discriminability of base classes and harm the transferability to novel classes, the negative margin technique has lower discriminability of base classes but can improve the feature transferability to novel classes.

\begin{table}[t]	
\centering
\addtolength{\tabcolsep}{2.5pt}
\scalebox{0.9}{
\begin{tabular}{cc cc cc}
\Xhline{2\arrayrulewidth}
\textbf{Label} & \textbf{Negative} & \multicolumn{2}{c}{\textbf{Base}} & \multicolumn{2}{c}{\textbf{Val}} \\
\textbf{Smoothing} & \textbf{Margin} & \textbf{1 shot} & \textbf{5 shot} & \textbf{1 shot} & \textbf{5 shot} \\
\hline 
           &            & 85.45 $\pm$ 0.67 & 93.72 $\pm$ 0.31 & 59.49 $\pm$ 0.90 & 78.90 $\pm$ 0.61 \\ 
\checkmark &            & \textbf{88.93 $\pm$ 0.62} & \textbf{93.93 $\pm$ 0.30} & 58.02 $\pm$ 0.97 & 75.87  $\pm$ 0.71 \\
           & \checkmark & 79.92 $\pm$ 0.81 & 92.34 $\pm$ 0.38 & \textbf{63.68 $\pm$ 0.86} & \textbf{81.60 $\pm$ 0.56} \\
\Xhline{2\arrayrulewidth}
\end{tabular}}
\caption{Comparison with label smoothing on the 5-way 1-shot and 5-shot accuracy for the base and validation classes in the mini-ImageNet dataset with ResNet-18 as backbone. For label smoothing, $\epsilon$ is set as 0.05. For negative margin, $m$ in pre-training stage is fixed to $-0.3$.}
\label{tab:label_smoothing}
\end{table}

\clearpage
{
\bibliographystyle{splncs04}
\bibliography{egbib}
}

\end{document}